\pdfoutput=1

\documentclass{article}

\PassOptionsToPackage{numbers, compress}{natbib}
\usepackage[preprint]{neurips_2020}

\usepackage[utf8]{inputenc} 
\usepackage[T1]{fontenc}    
\usepackage{url}            
\usepackage{booktabs}       
\usepackage{amsfonts}       
\usepackage{nicefrac}       
\usepackage{microtype}      

    \usepackage{graphicx}

    \usepackage{amsmath,amsfonts,amssymb}
    \usepackage{comment, textcomp, xcolor}
    \usepackage{ccaption, nicefrac, gensymb}
    \usepackage[labelfont=bf]{caption}
    
    \def\a{\textbf{a}}
    \def\b{\textbf{b}}
    \def\c{\textbf{c}}
    \def\d{\textbf{d}}
    \def\e{\textbf{e}}
    \def\f{\textbf{f}}

    \def\x{\mathbf x}
    \def\y{\mathbf y}

    \def\X{\mathbf X}

    \DeclareMathOperator*{\argmax}{arg\,max}
    \DeclareMathOperator*{\argmin}{arg\,min}

    \newcommand\inv[1]{#1\raisebox{1.15ex}{$\scriptscriptstyle-\!1$}}


\title{High-contrast ``gaudy'' images improve the training of deep neural network models of visual cortex}

\author{
  Benjamin R. Cowley, Jonathan W. Pillow\\
  Princeton Neuroscience Institute, Princeton University\\
  \texttt{\{bcowley, jpillow\}@princeton.edu}\\
}

\begin{document}

\maketitle

\begin{abstract}

A key challenge in understanding the sensory transformations of the visual system is to obtain a highly predictive model of responses from visual cortical neurons. Deep neural networks (DNNs) provide a promising candidate for such a model. However, DNNs require orders of magnitude more training data than neuroscientists can collect from real neurons because experimental recording time is severely limited. This motivates us to find images that train highly-predictive DNNs with as little training data as possible. We propose gaudy images---high-contrast binarized versions of natural images---to efficiently train DNNs. In extensive simulation experiments, we find that training DNNs with gaudy images substantially reduces the number of training images needed to accurately predict the simulated responses of visual cortical neurons. We also find that gaudy images, chosen \emph{before} training, outperform images chosen \emph{during} training by active learning algorithms. Thus, gaudy images overemphasize features of natural images, especially edges, that are the most important for efficiently training DNNs. We believe gaudy images will aid in the modeling of visual cortical neurons, potentially opening new scientific questions about visual processing, as well as aid general practitioners that seek ways to improve the training of DNNs.

\end{abstract}

\section*{Introduction}

A major goal in systems neuroscience is to understand the sensory transformations of the visual system \cite{heeger1996, dicarlo2012}. A key part of this goal is to find a model that accurately maps natural images to the responses of visual cortical neurons. To date, the most successful applications of this approach use visual features extracted from natural images by pre-trained models \cite{yamins2014, klindt2017, abbasi2018, sinz2018, schrimpf2018, cadena2019, kindel2019, zhang2019}. The basic setup is to pass an image as input into a DNN pre-trained for object recognition. One then maps activity of hidden units from a middle layer of this pre-trained DNN to neural responses (e.g., responses from neurons in monkey V4, a mid-level visual area, \cite{yamins2014}). This mapping is almost always chosen to be linear to avoid overfitting to the small amount of training data offered by neurophysiological experiments (where recording time is limited and costly). However, the true mapping between features and responses is likely nonlinear. Our goal is to train a more expressive mapping: a readout network (i.e., a DNN) that can fit to nonlinear mappings. Because DNNs tend to overfit with limited training data, we optimize images to train the readout network as accurately as possible with as little training data as possible (i.e., minimize recording time).

Here we report a surprising finding: We show that a simple manipulation of natural images substantially reduces the number of training images needed. This manipulation, inspired by active learning theory, maximizes the dynamic range of each input dimension in order to drive the activity of hidden units in the readout network as much as possible. It achieves this by setting the pixel intensity $p$ of each color channel to either 0 or 255, depending on if $p$ is less than or greater than the image's mean pixel intensity, respectively. We refer to the resulting images as ``gaudy'' images for their flashy bright colors and strong local contrasts. In extensive simulation experiments, we find that gaudy images improve the training of readout networks with different activation functions, different numbers of layers, different architectures, and different pre-trained DNNs used to simulate visual cortical responses. We find that gaudy images lead to large, informative gradients for backpropagation primarily by overemphasizing edges. In addition, we find that gaudy images, generated \emph{before} training, lead to performances on par with or greater than those achieved by images chosen \emph{during} training by active learning algorithms. This suggests that gaudy images overemphasize the features of natural images (e.g., high-contrast edges) most important to efficiently train the readout network---features that active learning algorithms must find without explicit direction.

Our results are likely to be of broad interest. Visual neuroscientists can test the efficacy of gaudy images to train models to predict visual cortical responses, and it is an open scientific question whether these neurons respond differently to gaudy images versus normal natural images. In addition, improving models of visual cortical neurons will improve methods that incorporate these models as a component, such as adaptive stimulus selection techniques to optimize neural responses \cite{benda2007, dimattina2011, cowley2017, ponce2019, walker2019, bashivan2019}. Researchers studying the similarities and differences of image representations between different DNNs trained for object recognition \cite{morcos2018, kornblith2019, lu2018} will also likely benefit from using gaudy images to probe these representations. In general, although we systematically test gaudy images for a particular regression setting with limited data, we suspect gaudy images will provide a useful training signal for transfer learning across many tasks, including classification tasks such as object recognition.

\section{Gaudy images improve the training of generalized linear models.}
        
What is the optimal set of images to train DNNs with as little training data as possible? To begin to answer this question, we recall a counterintuitive theoretical result from active learning and optimal experimental design: In the case of a linear relationship between stimulus and response, the optimal strategy for fitting a linear model is to choose stimuli \emph{before} collecting responses or training the model \cite{pukelsheim2006, jaakkola2006}---contrary to the strategy of the typical active learning algorithm, which chooses stimuli \emph{during} model training. The optimally chosen stimuli are outliers of the dataset (i.e., stimuli that maximize the variance of the input variables). We first give the mathematical underpinnings for this theory and then propose gaudy images as instances of such optimal stimuli.

Formally, let's assume a linear mapping between image $\x \in \mathbb{R}^K$ (for $K$ pixels) and response $y$: $y = \beta^T \x + \epsilon$, with weight vector $\beta \in \mathbb{R}^K$ and noise $\epsilon \sim \mathcal{N}(0,\sigma_\epsilon^2)$. We model this mapping with $\hat{y} = \hat{\beta}^T \x$, where $\hat{\beta} = \inv{\Sigma} \X \y$ for a set of (re-centered) training images $\X \in \mathcal{R}^{K \times N}$ (for $K$ pixels and $N$ images), unnormalized covariance matrix $\Sigma = \X \X^T$, and responses $\y \in \mathbb{R}^N$. Consider the expected error that represents how well our estimate $\hat{\beta}$ matches that of ground truth $\beta$: $E[\|\beta - \hat{\beta}\|^2_2]$. Given previously-shown images $\X$, we seek a new, unshown image $\x_\textrm{next}$ from the set of all possible (re-centered) images $\mathcal{X}$ that minimizes this error after training $\hat{\beta}$ on $\x_\textrm{next}$ and its response $y_\textrm{next}$. Importantly, we do not know $y_\textrm{next}$, so we cannot ask which image has the largest prediction error $\|y_\textrm{next} - \hat{y}(\x_\textrm{next})\|_2^2$. Instead, we choose the image $\x_{\textrm{next}}$ that minimizes the error between the true $\beta$ and our estimate $\hat{\beta}$ trained on images $[\X, \x_\textrm{next}] \in \mathbb{R}^{K \times (N+1)}$ (see the Appendix for full derivation):

\begin{equation}
    \x_{\textrm{next}} = 
    \argmin\limits_{\x \in \mathcal{X}} \textrm{E}\Big[\big\|\beta - \hat{\beta}\big\|_2^2 \mid \big[\X, \x\big] \Big] =
    \argmax\limits_{\x} \frac{\x^T (\inv{\Sigma})^2 \x}{1 + \x^T \inv{\Sigma} \x}
    \label{eqn:new_point}
\end{equation}

Examining this objective function reveals that the image $\x_{\textrm{next}}$ that minimizes the error is the one that maximizes its projection magnitude along the eigenvectors of $\inv{\Sigma}$ with the largest eigenvalues. An easier way to intuit this optimization is to assume that the $K$ pixels of $\x$ are uncorrelated (via a change of basis); i.e., the covariance matrix of pixel intensities $\Sigma$ is a diagonal matrix with on-diagonal entries $\Sigma_{k,k} = \sigma_k^2$. Under this assumption, the optimization becomes the following:

\begin{align*}
    \x_{\textrm{next}} = 
    \argmax\limits_{\x \in \mathcal{X}} \frac{\x^T (\inv{\Sigma})^2 \x}{1 + \x^T \inv{\Sigma} \x} \; \overset{\x_i \perp \x_j}{=} \;
    \argmax\limits_{\x} \sum\limits_{k=1}^{K} \nicefrac{\x_k^2}{\sigma_k^2}
\end{align*}

where $\x_i \perp \x_j$ indicate the $i$th and $j$th variables in $\x$ are uncorrelated for all $i$ and $j$, $i \neq j$. Intuitively, we seek the image $\x$ that most increases the variances along the dimensions in pixel space with small variance $\sigma_\textrm{small}^2$. We do this in order to increase the strength of the weakest signal $\sigma_\textrm{small}^2$ relative to noise $\sigma_\epsilon^2$ (i.e., increase the signal-to-noise ratio $\sigma_\textrm{small}^2 / \sigma_\epsilon^2$). Note that the optimization in Eqn.~\ref{eqn:new_point} does not depend on previous responses $\y$ nor the current model's weights $\hat{\beta}$. Thus, $\x_\textrm{next}$ can be chosen \emph{before} training the model.

We could choose a new image $\x_\textrm{next}$ from a large image dataset $\mathcal{X}$ based on Eqn.~\ref{eqn:new_point}. However, this requires taking the inverse of a large $K \times K$ matrix (difficult for computational reasons; $K$ is the number of pixels), and choosing from natural images does not harness the full dynamic range of pixel intensities (i.e., 0 to 255). Instead, we use the intuition of Eqn.~\ref{eqn:new_point} to \emph{synthesize} images such that each pixel's variance is maximized. We achieve this by taking a natural image and setting each pixel intensity $p$ to either the maximum value ($p=255$) or the minimum value ($p=0$) depending on if $p$ is above or below the mean pixel intensity of the image, respectively. We call the resulting images ``gaudy'' images for their bright, over-the-top colors (examples in Fig.~\ref{fig:sec4_dnn_results}\b). We confirm that gaudy images increase the variance of each pixel dimension and yield larger objective values in Eqn.~\ref{eqn:new_point} than those of normal images (Supp. Fig.~\ref{fig:suppfig1_glm_results}). 

\begin{figure}
    \centering
    \includegraphics[width=\textwidth]{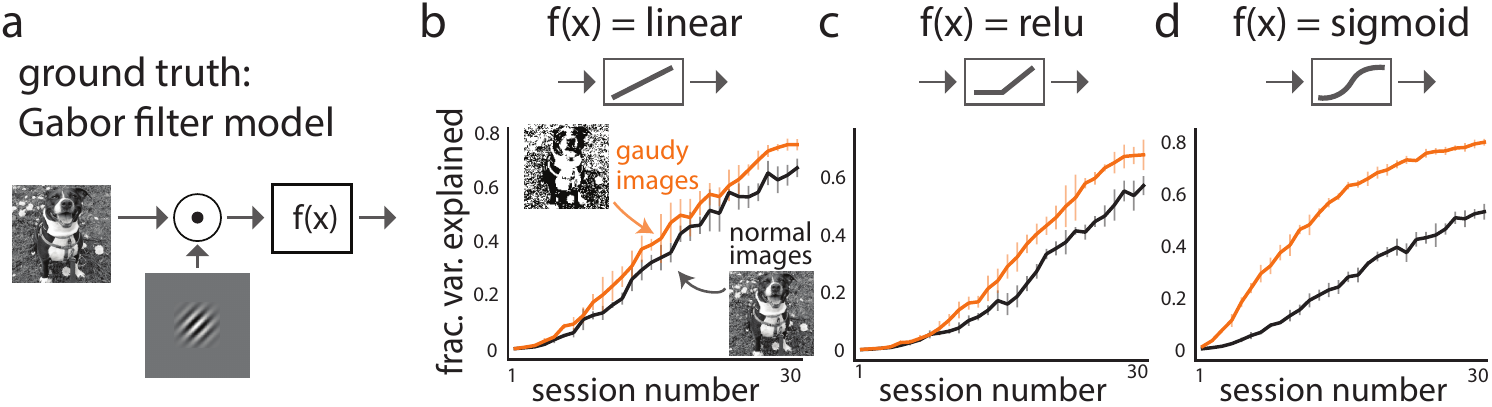}
    \caption{
    \textbf{Gaudy images improve the training of generalized linear models (GLMs).}
    \a. For ground truth, we use the Gabor filter model with activation function $f(x)$ with no noise added to the output (adding output noise yields similar results).
    \b. Training the filter weights of a GLM to predict ground truth responses. Fraction variance explained ($R^2$) is computed from responses to normal images only. The $f(x)$ for both GLM and ground truth is linear. Grayscale gaudy images have pixel intensities of either $0$ or $255$ (see text). 
    \c. Same as \b\ except $f(x)$ is a relu for both.
    \d. Same as \c\ except $f(x)$ is a sigmoid for both.
    Performance plateaus after \texttildelow50 sessions (Supp. Fig.~\ref{fig:suppfig1_glm_results}\d). 
    Error bars in \b-\d\ indicate 1~s.d. over 5~runs (some error bars are too small to see).}
    \label{fig:sec3_glm_results}
\end{figure}

As a first step, we test if gaudy images improve training for a model that satisfies the assumptions of the active learning theory (i.e., both model and ground truth mapping are linear). We simulate responses from a Gabor filter model (i.e., an instance of a generalized linear model or GLM). Each response is the result of a linear combination between an input grayscale image and the weights of a Gabor filter, which is then passed through an activation function $f(x)$ (Fig.~\ref{fig:sec3_glm_results}\a). We then train the filter weights of a GLM with the same activation function but with randomly-initialized filter weights and test the performance by predicting responses to heldout normal images (i.e., natural grayscale images, see Methods). We train the GLM over 30~sessions (500~images per session). 

We first test the setting in which the Gabor model and the GLM both have linear activation functions (upholding the assumption of Eqn.~\ref{eqn:new_point}). Performance gradually increases over sessions when trained on normal images (Fig.~\ref{fig:sec3_glm_results}b, black). As expected from the theory of Eqn.~\ref{eqn:new_point}, when we train the model on gaudy images, we see an increase in performance (Fig.~\ref{fig:sec3_glm_results}\b, orange line above black line). We note that this increase is small because a linear mapping requires little training data to achieve accurate predictions. 

Next, we consider the setting in which the Gabor model and the GLM both have the same nonlinear activation function. This nonlinearity breaks the assumption of Eqn.~\ref{eqn:new_point}, but we still find that gaudy images improve training for the relu activation function (Fig.~\ref{fig:sec3_glm_results}\c) and the sigmoid activation function (Fig.~\ref{fig:sec3_glm_results}\d). Interestingly, the effect of gaudy images on performance is enhanced as the activation function $f(x)$ becomes more heavily nonlinear (Fig.~\ref{fig:sec3_glm_results}, the differences between the orange and black lines increase from \b\ to \d). This is unexpected, as outliers (i.e., gaudy images) presumably lead to responses at the extremes of the activation functions (e.g., 0 or 1 for the sigmoid) where the derivatives are 0 (thus providing no gradient information). Instead, it appears gaudy images are better able to drive diverse outputs of the linear filter (especially for randomly-initialized weights that read out a random dimension in pixel space). This in turn allows error to more easily propagate to the weights of the linear filter. Overall, these results suggest that gaudy images may improve the training of DNNs, as DNNs are built up from layers of relu or sigmoid units like the ones here.

\section{Gaudy images improve the training of deep neural networks.}

\begin{figure}
    \centering
    \includegraphics[width=\textwidth]{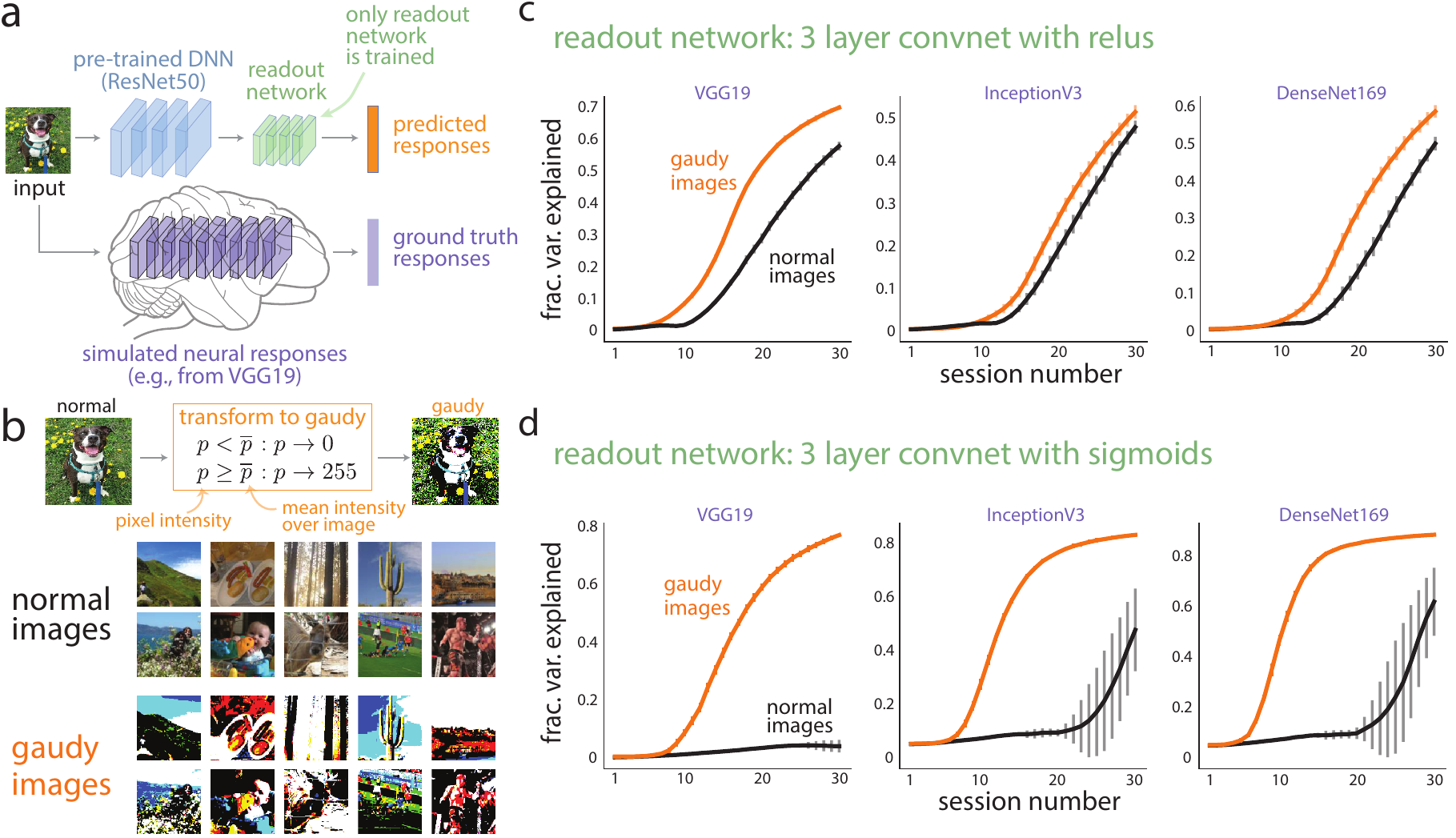}
    \caption{
    \textbf{Gaudy images improve the training of DNNs.}
    \a. Our simulation setup. We simulate visual cortical responses with responses of hidden units from a middle layer of a DNN pre-trained for object recognition (purple). We predict these ``ground truth'' responses using a readout network (green) to map the features from a middle layer of a different pre-trained DNN (blue) to predicted responses (orange). Our goal is to train the readout network (all other DNNs are fixed).
    \b. Example normal images (top 2~rows) are less bright and have less contrast versus their gaudy versions (bottom 2~rows). 
    \c. Results for predicting ground truth responses to normal images of different pre-trained DNNs (VGG19, InceptionV3, and DenseNet169). The readout network has three convolutional layers with relu activation functions. Performance plateaus after \texttildelow50 sessions (Supp. Fig.~\ref{fig:suppfig3_dnn_results_fracvars}\a).
    \d. Same as \c\ except for sigmoid activation functions. Error bars in panels of \c\ and \d\ indicate 1~s.d. for 5~runs (some error bars are too small to see).
    }
    \label{fig:sec4_dnn_results}
\end{figure}

Given that gaudy images improve training for GLMs (Fig.~\ref{fig:sec3_glm_results}), we next ask if gaudy images improve the training of DNNs, which are essentially feedforward stacks of GLMs. To do so, we create a realistic simulation paradigm that mimics the data-limited regression problem faced by visual neuroscientists seeking to characterize neural responses from natural images (Fig.~\ref{fig:sec4_dnn_results}\a). We first describe this paradigm, and then justify the simulation's realism to biological experiments. 

To simulate visual cortical responses from a mid-level visual area (e.g., monkey V4), we consider the responses of 100~hidden units (or `neurons') from a middle layer of a DNN pre-trained for object recognition (Fig.~\ref{fig:sec4_dnn_results}\a, purple, see Methods for how the hidden units were chosen). Our goal is to predict these `ground truth' responses. To do this, we employ transfer learning. We first pass an image as input into ResNet50 \cite{he2016}, a different pre-trained DNN from the ground truth DNN. We take as features the activity of a middle layer (Fig.~\ref{fig:sec4_dnn_results}\a, blue) and then feed these features as input into a readout network (Fig.~\ref{fig:sec4_dnn_results}\a, green) which in turn outputs a vector of 100~predicted responses (Fig.~\ref{fig:sec4_dnn_results}\a, orange). Our goal is to train the readout network (with all pre-trained DNNs held fixed---no fine tuning) to predict ground truth responses as accurately as possible with as few training images as possible.

Current models of visual cortical neurons map the activity from a middle layer of a pre-trained DNN to neural responses \cite{yamins2016}. This mapping is almost always linear to avoid overfitting to the limited amount of experimental data. Here, we relax this assumption by using a DNN (i.e., the readout network) to fit nonlinear mappings. We find that with enough training data, a DNN (i.e., a nonlinear mapping) outperforms a linear mapping in this setting (see below).

We have designed our simulation setup and training procedure to realistically mimic neurophysiological experiments. The specific pre-trained DNNs we use are currently the most predictive models of visual cortical responses in monkey V4, a mid-level visual area \cite{schrimpf2018}, and transfer learning is commonly employed to predict visual cortical responses \cite{yamins2016}. In standard neurophysiological experiments, neural activity of \texttildelow100~neurons is recorded during a session lasting \texttildelow2~hours per day, during which a set of \texttildelow500 unique images are presented to the animal \cite[e.g.,][]{cowley2017}. Experiments last for \texttildelow30~days. To mimic this setting, we train the readout network for 30~sessions with 500~images per session. We assume the same neurons are recorded across sessions, which is experimentally possible with calicum imaging \cite{ji2016, sheintuch2017} or neural stitching techniques \cite{nonnenmacher2017, degenhart2020}. 

Given this setup, we now ask whether training on gaudy images increases the performance of the readout network more than that achieved by training on colorful, natural images (i.e., ``normal'' images). We compute a colorful gaudy image by setting each pixel intensity $p$ of a normal image to $p=0$ if $p$ is less than the mean pixel intensity of the image (taken over all RGB channels), and set $p=255$ otherwise. The resulting gaudy images have at most eight different colors, but still retain a surprising amount of information about the original image (Fig.~\ref{fig:sec4_dnn_results}\b). 

We consider a readout network with 3~convolutional layers and relu activation functions (see Methods for architecture details of all networks presented in this paper). We measure test performance on heldout normal images (fraction variance explained averaged over neurons, see Supp. Fig.~\ref{fig:suppfig2_dnn_results_sweep}\a\ and Methods). We train on 500~images per session. Instead of training on 500~gaudy images, we instead train on a mix of 250~gaudy images and 250~normal images, as this leads to higher performance (Supp. Fig.~\ref{fig:suppfig2_dnn_results_sweep}\b). This is because training on only outlying images (i.e., gaudy images) makes the training set and test set distributions too different. 
We find that training on gaudy images increases performance, sometimes substantially, versus training on normal images (Fig.~\ref{fig:sec4_dnn_results}\c, orange above black lines). This result holds across different pre-trained DNNs used for surrogate responses, including VGG19 \cite{simonyan2014}, InceptionV3 \cite{szegedy2016}, and DenseNet169 \cite{huang2017}. Switching the roles of ResNet50 with one of these DNNs leads to similar results. When we change all activation functions to sigmoid, gaudy images again improve training over that of normal images (Fig.~\ref{fig:sec4_dnn_results}\d, orange above black lines). Interestingly, we observe larger increases in performance for the sigmoid network than for the relu network, similar to the observed boosts in performance for the sigmoid GLM over the relu GLM (Fig.~\ref{fig:sec3_glm_results}\c\ and \d). This suggests that gaudy images are more effective at training DNNs with stronger nonlinear activation functions. We also find that gaudy images improve training for readout networks with large numbers of layers (e.g., 10~layers, Supp. Fig.~\ref{fig:suppfig2_dnn_results_sweep}\c) and for a readout network with an architecture that comprises ResNet-blocks (Supp. Fig.~\ref{fig:suppfig3_dnn_results_fracvars}\b). We confirm that these readout networks (i.e., nonlinear mappings) outperform a linear mapping; however, training a linear mapping with gaudy images does not outperform training with normal images (Supp. Fig.~\ref{fig:suppfig3_dnn_results_fracvars}\c). This is expected because this setting breaks the assumption of the theory in Eqn.~\ref{eqn:new_point} that the ground truth mapping must also be linear (Supp. Fig.~\ref{fig:suppfig3_dnn_results_fracvars}\c). Overall, our results indicate that gaudy images substantially improve the training of DNNs to predict visual cortical responses.

\section{Gaudy images improve training by increasing the contrast of edges.}

\begin{figure}
    \centering
    \includegraphics[width=\textwidth]{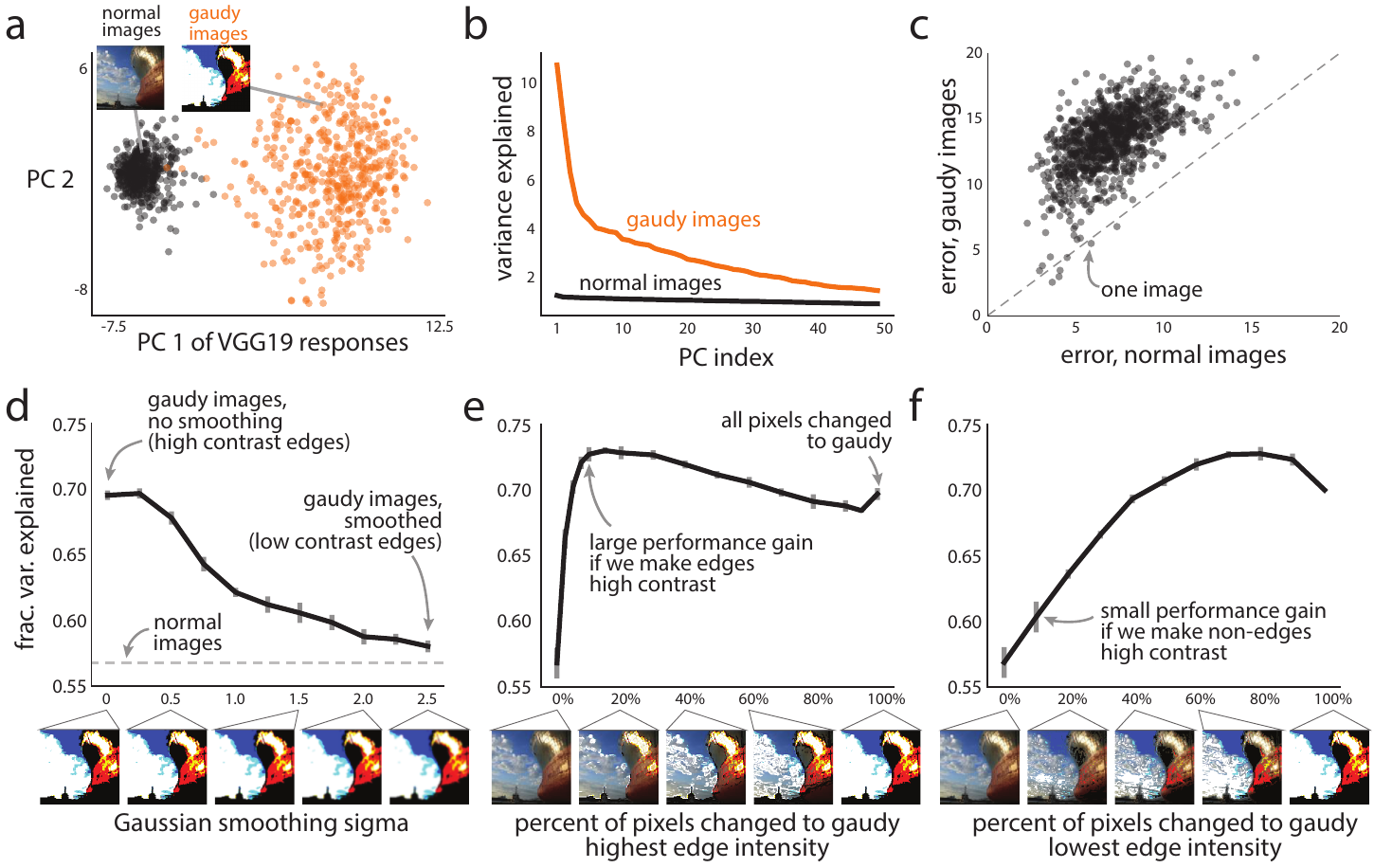}
    \caption{
    \textbf{Gaudy images improve training primarily by driving diverse responses and increasing the contrast of edges.}
    \a. The top two principal components (PCs) of VGG19 responses to normal images (black dots) and their gaudy versions (orange dots).
    \b. The variance of VGG19 responses for each of the top 50~PCs, where PCA is applied 
        separately to responses from either 5,000~normal images (black) or their gaudy versions (orange).  
    \c. Error between VGG19 responses and predicted responses of a trained readout network (3-layers with relus) to heldout images (either normal or its gaudy version).
    \d. Performance for training on gaudy images smoothed with a Gaussian filter. A Gaussian smoothing sigma of 1.0 corresponds to a s.d. of 1~pixel. 
    \e. We transform pixels with the highest edge intensities to gaudy (choosing the edge intensity threshold based on percentage quantiles). At 100\%, all pixels are transformed to gaudy.
    \f. Same as in \e, except that we transform pixels with the lowest edge intensities to gaudy. 
    In \d-\f, reported fraction explained variances are from the 30th session of a trained readout network (3~layers with relus) to predict VGG19 responses to heldout normal images.
    Error bars in panels \d-\f\ indicate 1~s.d. over 5~runs (some error bars are too small to see).
    }
    \label{fig:sec5_why_gaudi}
\end{figure}

The impressive training improvement of gaudy images begs the question: What makes gaudy images so special? From an optimization standpoint, gaudy images likely have two advantages. First, they drive surrogate responses to regions in response space not reachable by normal images. For example, VGG19 responses to gaudy images reside in regions far from those to normal images (Fig.~\ref{fig:sec5_why_gaudi}\a, orange dots far from black dots), and responses to gaudy images are more diverse along many response dimensions (Fig.~\ref{fig:sec5_why_gaudi}\b, orange line above black line). The diverse responses of gaudy images also lead to larger prediction errors (Fig.~\ref{fig:sec5_why_gaudi}\c), which in turn lead to larger and more informative gradients that more strongly impact the weights of earlier layers. The second advantage is that gaudy images lead to more diverse activations in the readout network than those from normal images, even for untrained, random networks (Supp. Fig.~\ref{fig:suppfig4_why_gaudy}\a). This is important because a hidden unit must have its input vary in order to fit the variations of some desired output. Larger variations of this input likely lead to better fits to the variations of the output. Taken together, gaudy images improve optimization by enlarging the magnitude of the gradient and more strongly varying the inputs.

An important feature of gaudy images is their high contrast---performing the gaudy transformation is akin to substantially increasing an image's contrast (by 400\%, Supp. Fig.~\ref{fig:suppfig4_why_gaudy}\b). However, we find a more parsimonious feature in gaudy images that better explains their ability to efficiently train DNNs: Gaudy images overemphasize edges. This intuitively follows from the idea that high-contrast edges strongly drive the edge-detectors of early DNN layers, which in turn more strongly drive feature detectors of later DNN layers. Next, we show that these high-contrast edges are necessary and sufficient to efficiently train DNNs.

To test for necessity, we smooth the gaudy images (i.e., decreasing the contrast of edges), and find that performance decreases for smoother images (Fig.~\ref{fig:sec5_why_gaudi}\d). Thus, high-contrast edges are necessary to increase performance. Next, we ask if high-contrast edges are sufficient to increase performance. To test this, we perform edge detection on each image and compute an edge intensity for each pixel (defined as the norm of the x- and y-gradients using a Sobel edge-detecting filter). Starting with the original image (Fig.~\ref{fig:sec5_why_gaudi}\e, 0\%), we transform a percentage of pixels with the highest edge intensity to gaudy, leaving the remaining pixels unchanged. This increases the contrast of edges (Fig.~\ref{fig:sec5_why_gaudi}\e, compare the clouds and background's silhouette between 0\%\ and 20\%\ image insets). Surprisingly, we find that we need to change only 10\% of the pixels with the highest edge intensities to gaudy to achieve a similar (and even larger) performance than that of changing all pixels to gaudy (Fig.~\ref{fig:sec5_why_gaudi}\e, compare 10\%\ to 100\%). On the other hand, transforming pixels with the lowest edge intensities to gaudy does not lead to such increases in performance (Fig.~\ref{fig:sec5_why_gaudi}\f, 10\%). Thus, high contrast edges are sufficient to improve training. Interestingly, transforming 60\%-80\%\ of low-edge-intensity pixels still yields an increase in performance above that of 100\%\ (Fig.~\ref{fig:sec5_why_gaudi}\f, 60\%\ to 80\%), likely because the images for 60\%\ to 80\%\ have more high-contrast edges than those for 100\%\ (Fig.~\ref{fig:sec5_why_gaudi}\f, 60\%\ vs. 100\%\ image insets). In additional analyses, we find that removing texture (but keeping high-contrast edges) leads to high performance (Supp. Fig.~\ref{fig:suppfig4_why_gaudy}\c), while altering color statistics does not achieve the same level of performance as gaudy images (Supp. Fig.~\ref{fig:suppfig4_why_gaudy}\d). Overall, these results suggest that high-contrast edges are both necessary and sufficient to improve the training of DNNs.

\section{Gaudy images train DNNs better than active learning.}

\begin{figure}
    \centering
    \includegraphics[width=\textwidth]{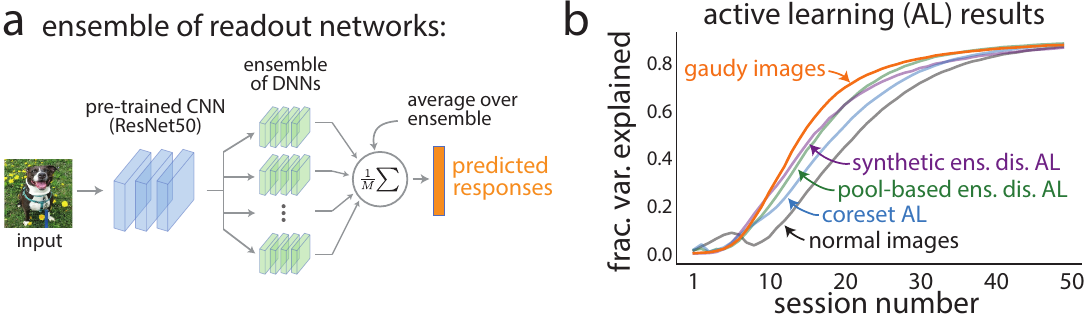}
    \caption{
    \textbf{Gaudy images, chosen \emph{before} training, improve performance more than that of images chosen \emph{during} training by active learning.}
    \a. Our simulation setup for an ensemble of readout networks (see text for details).
    \b. For each session, we train the ensemble of networks with normal images, gaudy images, or images chosen by active learning (AL) algorithms (described in text). The initial ``bump'' for normal images likely arises from an initial local optimum that quickly overfits; the bump is present for a range of learning rates. Example chosen/synthesized images as well as error bars can be found in Supplemental Figure~\ref{fig:suppfig5_AL_ensemble_results}.
    }
    \label{fig:sec6_AL_results}
\end{figure}

We generate gaudy images \emph{before} training. However, it might be the case that adaptively choosing images \emph{during} training (e.g., choosing images based on the model's current uncertainty) will increase performance even more. These adaptive approaches, known as active learning (AL) algorithms \cite{settles2012}, search the space of image statistics (a subset of which is gaudy-like statistics) to find the features of natural images most important for training. Unexpectedly, we find that gaudy images yield similar and sometimes even larger gains in performance than those from AL algorithms that either access a large number of normal images or synthesize normal images.

To test the performance of AL algorithms versus gaudy images, we employ a model for the readout network conducive for AL. The model comprises an ensemble of DNNs, each with the same network architecture but different initial random weights (Fig.~\ref{fig:sec6_AL_results}\a, green). Each ensemble DNN is trained separately but the outputs are averaged across the ensemble to compute the predicted responses. Consistent with other ensemble approaches for deep learning \cite{huang2017snapshot}, we find that an ensemble of DNNs yields higher performance than a single DNN (Supp. Fig.~\ref{fig:suppfig5_AL_ensemble_results}\a). The intuition for this is that a single DNN likely overfits to some regions of response space, leading to incorrect interpolations. However, because each ensemble DNN is initialized differently and trained separately, each ensemble DNN likely overfits in a different way, and these differences are averaged away over the ensemble.

While AL algorithms to train DNNs have been proposed for object recognition tasks \cite{tran2019, sinha2019, sener2017, beluch2018, gal2017, kirsch2019, yoo2019, ducoffe2018, munjal2020, ash2019}, few exist to train DNNs for regression tasks, which require a different notion of uncertainty than that of classification tasks \cite{tsymbalov2018}. Here, we propose three different AL algorithms based on two state-of-the-art AL algorithms for object recognition \cite{sener2017, beluch2018}. The first two rely on a notion of uncertainty, defined as the disagreement over the ensemble DNNs. We find that images with the largest ensemble disagreement also have the largest error (Supp. Fig.~\ref{fig:suppfig5_AL_ensemble_results}\b), suggesting these images will better guide the next gradient step versus randomly-chosen images. The first algorithm (`pool-based ens. dis. AL') is an extension of ensemble approaches for AL \cite{beluch2018, gal2017, tsymbalov2018} and chooses images with the largest ensemble disagreement from a large pool of normal images. The second algorithm (`synthetic ens. dis. AL') synthesizes images from a natural prior of normal images to maximize ensemble disagreement. The natural prior is in the form of a generator network trained in a GAN-like fashion \cite{dosovitskiy2016}. The final algorithm (`coreset AL'), extended from previous work \cite{sener2017}, does not maximize ensemble disagreement but rather maximizes the diversity of responses between previously-chosen images and candidate images chosen from a large pool via a coreset approach. Example images chosen by these AL algorithms are presented in Supplemental Figure~\ref{fig:suppfig5_AL_ensemble_results} and further details about these algorithms are in Methods.

We now compare the performance of gaudy images to that of various active learning algorithms. As found in the previous section, training on gaudy images yields a larger performance than that of normal images (Fig.~\ref{fig:sec6_AL_results}\b, orange line above black line). The reported performances here are larger than those for a single DNN (Fig.~\ref{fig:sec4_dnn_results}\c, left panel) because here we use an ensemble of DNNs. In addition, we find that gaudy images yield larger improvements than any active learning algorithm (Fig.~\ref{fig:sec6_AL_results}\b, orange line above purple, green, and blue lines). We find similar results when predicting simulated neural responses from other pre-trained DNNs (Supp. Fig.~\ref{fig:suppfig5_AL_ensemble_results}\c, performance gains for gaudy images are on par with or larger than those for active learning algorithms). These results indicate that gaudy images, chosen \emph{before} training, lead to performances similar to or even greater than those from images chosen adaptively \emph{during} training. This suggests that gaudy images overemphasize features of natural images (e.g., high-contrast edges) that are the most beneficial for training DNNs in this setting (else the active learning algorithms would have identified other features to increase performance).

\section*{Discussion}
        
We have proposed gaudy images to efficiently train DNNs to predict the responses of visual cortical neurons from features of natural images. Through extensive simulations, we have found that gaudy images improve training for all tested DNN architectures and activation functions. These improvements arise from the high-contrast edges overemphasized in gaudy images. Our motivation to use gaudy images has come from the counterintuitive theoretical result that, under certain assumptions, the optimal active learning strategy is to choose the most outlying inputs before any training. Surprisingly, after relaxing those assumptions, we still have found gaudy images, computed \emph{before} training, outperform active learning algorithms that choose or synthesize images \emph{during} training. These results suggest that gaudy images constitute an important ingredient to train DNNs.

We have tested gaudy images on somewhat small DNNs (i.e., less than 2~million parameters), and it remains an open question if gaudy images are beneficial for training larger DNNs to perform classification tasks (e.g., training ResNet on ImageNet) or for unsupervised tasks (e.g., training a generative adversarial network). We believe gaudy images will be most helpful in tasks that involve transfer learning (i.e., using visual features from a DNN pre-trained for object recognition), as gaudy images likely induce too large a mismatch between training and test set distributions when training a large DNN that takes pixel intensities as input. For classification tasks, gaudy images may be a useful transformation for data augmention methods \cite{sohn2020, perez2017, shorten2019}, which already (weakly) transform the contrast of images. Because data augmentation assumes that class labels are invariant to weak image transformations, the class labels between a normal image and its gaudy version ideally would not be largely different. Indeed, we find this to be the case (Supp. Fig.~\ref{fig:suppfig6_classification_results}). Thus, even when we do not have the true labels for gaudy images, we may still benefit from using gaudy images as one of the transformations performed in data augmentation.

Our work follows a long line of previous studies in computational neuroscience that perform extensive, realistic simulations as a proof-of-concept \cite{moreno2014, parthasarathy2017, xiao2019}, including studies that train neural response models with active learning \cite{dimattina2011, lewi2009, park2011}, before performing costly biological experiments to test their predictions \cite{ponce2019, sharpee2013, montijn2019, brackbill2020}. Because our simulated neural responses come from models predictive of visual cortical responses  \citep{schrimpf2018}, we suspect our results will likely carry over to neuroscientific experiments. However, it is an open scientific question whether visual cortical neurons have a similar preference for gaudy images as that of DNN neurons. Indeed, how does a visual cortical neuron respond to a normal image, a gaudy version of the same image, or a version in which all edges are high contrast (i.e., transforming only 10\%\ of all pixels, Fig.~\ref{fig:sec5_why_gaudi}\e)? As we learn more about the structure of natural images relevant to training DNNs, we may better identify the priors that the visual system uses to extract useful information from natural images.

\section*{Broader Impact}
The goal of our work is to train DNNs as accurately as possible with as little training data as possible. This reduces the amount of research hours needed to collect data (e.g., an experimenter collecting neural data) as well as reduces the compute time for training DNNs, which in turn reduces the amount of C02 emitted. We focus on a specific regression problem of interest in computational neuroscience versus object recognition, for which many active learning algorithms already exist and would likely require tens of thousands of hours of GPU compute time to perform our analyses. All of our work was performed on a small cluster of eight 12-Gb GPUs (GeForce RTX 2080 Ti). We estimate that \texttildelow8,000~GPU hours in total were used, emitting \texttildelow200~lbs of CO2 (assuming 40~GPU hours consumes 10~kWh). This is equivalent to driving \texttildelow230~miles in a car. We do not foresee any short-term negative consequences to society from our work. Code to produce the figures in this paper is available at [link removed until published].

\section*{Acknowledgments}
B.R.C. was supported by a CV Starr Fellowship. J.W.P. was supported by grants from the Simons Collaboration on the Global Brain (SCGB AWD543027) and the NIH BRAIN initiative (NS104899 and R01EB026946).


\clearpage

\section*{Methods}

This section describes the image dataset, simulations, and network architectures used in our work. Code to produce the figures in this paper is available at [link removed until published]. Our simulations were coded in Python, using Keras \cite{chollet2015} and TensorFlow \cite{abadi2016}.

\section*{Image dataset}

    For our image dataset, we randomly sample 12~million ``natural'' colorful images from the Yahoo Flickr Creative Commons 100 Million Dataset (YFCC100M) \citep{thomee2016}, which contains \texttildelow100~million images uploaded by users to Flickr between 2004 and 2014. Images are unlabeled (i.e., no content information), and need not contain a recognizable object. We resize each RGB image to $112\times112$~pixels, randomly cropping the image to have the same number of row and column pixels. We choose this dataset primarily to ensure that our training images are different from the images used to train the chosen pre-trained DNNs (i.e., from ImageNet \cite{deng2009}).

\section*{Simulations with generalized linear models}
            
    We test to what extent gaudy images improve the training of generalized linear models (GLMs) when the ground truth model is also a GLM. We grayscale each image (112 $\times$ 112 pixels) by averaging over the RGB channels for each pixel. For the ground truth model (Fig.~\ref{fig:sec3_glm_results}\a), we use a $112 \times 112$ Gabor filter with spatial frequency 0.1 cycles/pixel, bandwidth 0.5, orientation angle 45\degree, and location at the center of the image. The output response (a single variable) is computed by taking a dot product between the input image and the Gabor filter, which is then passed through an activation function (either linear, relu, or sigmoid). We do not add noise to the output, as we already see training improvements without noise; however, adding output noise leads to similar improvements when training on gaudy images. For the sigmoid, we first normalized the dot product (dividing by a constant factor equal to 1,000) before passing it through the sigmoid to ensure responses are within a reasonable range (i.e., not simply 0 or 1). 
    
    To predict ground truth responses, we consider GLMs with three different activation functions: linear, relu, and sigmoid. We train the randomly-initialized $112 \times 112$ filter weights of each GLM. We only train a GLM to predict ground truth responses from the Gabor filter model with the \emph{same} activation function. This is to ensure that we uphold the assumption made by the active learning theory in Eqn.~\ref{eqn:new_point} (i.e., fitting a linear mapping to a ground truth linear function). We train GLMs with re-centered images (i.e., we subtract 110 from each pixel intensity) for 30~``sessions''. Each session comprises 500~training images, and GLMs are trained with five passes (or epochs) over the 500~images for that session. To measure test performance, we compute the fraction of variance explained (also known as the coefficient of determination) for 4,000~heldout images. For training optimization, we use SGD with momentum, where the learning rates and momenta are chosen such that the fraction variance explained is roughly 0.5 for GLMs trained with normal images after 30~sessions. The learning rates are 1e-7, 3e-7, 1e-7 and the momenta are 0.99, 0.99, and 0.7 for linear, relu, and sigmoid activation functions respectively. Other reasonable learning rates and momentum terms yield similar results. The batch size is 64 images. For each session, we either train on 500~``normal'' images (i.e., grayscale, re-centered images) or 500~``gaudy'' images. To compute a gaudy image, we take a normal image and set each pixel intensity $p$ to $p=0$ if $p < \bar{p}$, else $p=255$, where $\bar{p}$ is the mean pixel intensity over all pixels of the image.

\section*{Simulations with deep neural networks}

    We seek to test if gaudy images improve the training of DNNs to predict the responses of visual cortical neurons to natural images. Ideally, we would predict these responses by training a large DNN with tens of millions of parameters from scratch, end-to-end. However, this would require recording responses to hundreds of thousands of images, currently not possible with the severely limited amount of recording time in neuroscientific experiments. Instead, to model these responses, we employ transfer learning, commonly used by other studies \cite{yamins2014, schrimpf2018, abbasi2018, cadena2019}. We pass the input image through the early layers of a DNN pre-trained for object recognition. This pre-trained DNN outputs a set of features, which we then feed as input into a readout network. The readout network in turn outputs the predicted responses. Our goal is to train the weights of the readout network (keeping the pre-trained DNN fixed) as accurately as possible with as little training data as possible.
    
    To simulate responses of visual cortical neurons, we use the responses of hidden units from middle layers of DNNs (pre-trained on object recognition, different from the pre-trained DNN used to extract features). Previous studies have shown that the same pre-trained DNNs we use in this paper predict roughly 60\%\ of the variance of visual cortical responses in macaque V4 \cite{schrimpf2018}, suggesting the results of our simulations will likely carry over to neuroscientific experiments.

    \subsection*{Simulated neural responses: Pre-trained DNN responses}
        We consider three different pre-trained DNNs (all trained for object recognition on ImageNet) to simulate responses of visual cortical neurons. We choose a middle layer of each DNN, and perform average pooling ($2 \times 2$ pooling window) to reduce the number of variables. We choose a middle layer (versus an early or late layer), as the image representations of these layers best match that of visual cortical neurons in monkey V4 \cite{schrimpf2018}. Here we mention the specific layers we choose; however, choosing other middle layers leads to similar results. The pre-trained DNNs are as follows:
        
        \begin{itemize}
            \item VGG19 \cite{simonyan2014}: A popular DNN known for its relatively small convolutional filters ($3 \times 3$). We take the responses from layer \texttt{block4\_pool}.
            \item InceptionV3 \cite{szegedy2016}: A DNN that considers multiple kernels (e.g., $1 \times 1$, $3 \times 3$, and $5 \times 5$) at each stage, where the network's width and depth are balanced. We take the responses from layer \texttt{mixed4}.
            \item DenseNet169 \cite{huang2017}: A DNN where each layer has access to the outputs of all previous layers in an effort to avoid the vanishing gradient problem. We take the responses from layer \texttt{pool3\_pool}.
        \end{itemize}

        To match the typical number of visual cortical neurons recorded in a neuroscientific experiment (\texttildelow100~neurons), we consider responses from 100~hidden units from the same middle layer. Because a randomly-chosen hidden unit likely encodes little stimulus information (as up to 90\%\ of a network's weights can be pruned \cite{han2015, frankle2018}), we assume that hidden units with the largest response variance carry the most stimulus information. To this end, we apply PCA to the DNN middle layer's responses to 5,000~normal images and take the responses of the top 100~PCs as the simulate responses. Note that each response is a linear combination of the responses of all hidden units from the chosen middle layer (typically \texttildelow10,000~hidden units or greater). We scale the weights of each linear combination to ensure the responses of each output neuron have a mean of 0 and a standard deviation of 1.

    \subsection*{Model: Transfer learning and readout network}

        We employ transfer learning for our model that maps images (i.e., raw pixel intensities) to visual cortical responses. It consists of a sub-network of a pre-trained DNN (fixed and not trained) followed by a readout network. The sub-network comprises the early layers of ResNet50 \cite{he2016}, which leverages skip connections to increase the number of layers. For the output features of this sub-network, we perform average pooling ($2 \times 2$ pooling window) for the activations of middle layer \texttt{activation\_27}, yielding 50,176 feature variables of shape $7 \times 7 \times 1,024$ (i.e., an activity map of $7 \times 7$ for 1,024 filters). We then consider two readout networks with the same architecture but different activation functions (more network architectures tested in Supp. Fig.~\ref{fig:suppfig3_dnn_results_fracvars}). Each readout network comprises an initial convolutional layer ($1 \times 1$ kernel, linear activation function, 512~filters), which reduces the number of input variables, followed by 3~convolutional layers ($3 \times 3$ kernel, separable convolution, 512~filters) with either a relu activation function (Fig.~\ref{fig:sec4_dnn_results}\c) or sigmoid activation function (Fig.~\ref{fig:sec4_dnn_results}\d). The final layer is a linear, dense layer of 100~filters (to match the number of ground truth DNN neurons, 100). We use batch normalization \cite{ioffe2015} in every layer; without batch normalization, we found it difficult to train DNNs even with a large number of training images (e.g., 50,000~images).

    \subsection*{Training of readout networks}
    
        We initialize each readout network randomly using the default Keras settings, including the random sampling of initial weights from a Glorot uniform distribution. We train networks over 30~sessions, where each session consists of either 500~normal images or a mix of 250~normal images and 250~gaudy images (the latter yielding better performance than training on the full 500~gaudy images per session, Supp. Fig.\ref{fig:suppfig2_dnn_results_sweep}\b). The weights of the trained network from the previous session are carried over to be the initial weights of the network for the current session, and we train the network for 5~passes of the 500~images. We find that re-training a randomly-initialized network each session on data from all previous and current sessions leads to similar results. We optimize with SGD, using a momentum of 0.7 and batch size of 64. We choose learning rates to ensure the fraction variance explained for each network trained on normal images reaches \texttildelow0.5 after 30~sessions, although other reasonable learning rates yield similar results. The learning rates for the relu readout network are 1e-1, 8e-2, and 7e-2 for ground truth responses of VGG19, InceptionV3, and DenseNet169, respectively. The learning rates for the sigmoid network are 5e-1, 1e0, and 1e0, respectively. We test performance by computing the fraction variance explained ($R^2$) of ground truth responses to 4,000~normal images, computed for each neuron and then averaged over neurons.
        
        The colorful images are pre-processed and re-sized according to the required RGB format of the respective pre-trained DNN. To compute gaudy images, we set each pixel intensity $p$ to $p=0$ if $p < \bar{p}$, else $p=255$, where $\bar{p}$ is the mean pixel intensity over all channels and pixels of the image.

\section*{Simulations with active learning}

    We ask to what extent gaudy images, chosen \emph{before} training, lead to an increase performance versus that achieved by images chosen by active learning algorithms \emph{during} training. Inspired by the state-of-the-art active learning algorithms for deep neural networks performing object recognition \cite{sener2017, beluch2018}, we propose three new active learning algorithms for deep regression. These active learning algorithms choose or synthesize new images that maximize the uncertainty of the model or the diversity of the responses. We use an ensemble of DNNs for our model (Fig.~\ref{fig:sec6_AL_results}\a) in order to have a measure of uncertainty for active learning. We randomly initialize each ensemble DNN with a different seed and train each ensemble DNN separately with a different shuffled ordering of training samples. The predicted response of each neuron is the average predicted responses across ensemble DNNs for that neuron. We find that 25~ensemble DNNs yield the best performance (Supp. Fig. \ref{fig:suppfig5_AL_ensemble_results}\a).
    
    We propose three different active learning algorithms. The algorithms use batch learning (i.e., query the responses of chosen images in batches) as opposed to sequentially querying each image. Batch learning is a good fit for neuroscientific experiments, which require time-consuming pre-processing of neural activity (i.e., extracting spike counts from voltage traces or calcium imaging) that is difficult to perform in real-time. For each session, we train with a batch of 500~images, either 500~normal images or a mix of 250~normal images and 250~`non-normal' images (i.e., either gaudy images or images chosen by active learning). We use this mix because we find it leads to larger gains in performance (Supp. Fig.~\ref{fig:suppfig2_dnn_results_sweep}\b). The active learning algorithms are listed below. We provide example images either chosen or synthesized by these algorithms in Supplemental Figure~\ref{fig:suppfig5_AL_ensemble_results}\d.

    \begin{itemize}
        \item \emph{Pool-based ens. dis. AL} chooses 250~images from a large pool of 80,000~normal images. For each session, a new pool of 80,000~images are randomly chosen from the FlickR image dataset of 12~million images. For each candidate image, we compute a measure of ensemble disagreement (ens. dis.), defined as the median distance between predicted responses of the ensemble DNNs. We find taking the median performs better than computing the variance across ensemble DNNs (summed across neurons), likely because estimates of variance are noisy for a high-dimensional response space (here, 100~dimensions) with few samples (here, 25~ensemble DNNs). Because we choose images in batches, it might be the case that the top 250~candidate images with the highest ensemble disagreement drive similar responses. To account for this possibility, we take the top 2,000~candidate images with the highest ensemble disagreement, and pass these images through the ResNet50 sub-network to obtain their corresponding feature vectors. We then perform a coreset on these feature vectors. The coreset returns 250~of the 2,000~candidate images whose feature vectors are as far as possible from all other feature vectors in feature space. The images returned by this coreset are the ones used to train the ensemble model for the next session.
        
        \item \emph{coreset AL} is an extension of a coreset active learning algorithm for DNNs to perform object recognition \cite{sener2017}. For each session, we choose images from a large pool of 80,000~candidate normal images. To begin, we compute the minimum Euclidean distance for each candidate image between the vector of predicted responses to the candidate image and the vectors of predicted responses to all previously-trained images. We then choose the candidate image that has the largest of these Euclidean distances, and append that image to the coreset. We repeat this process 250~times to retrieve 250~chosen images. Our algorithm differs from the previous coreset algorithm \cite{sener2017} by performing coreset on the final output versus the penultimate layer as well training on a mix of 250~normal images and 250~chosen images versus training only on chosen images.

        \item \emph{Synthetic ens. dis. AL} synthesizes an image based on a natural prior of normal images to maximize the ensemble disagreement. Ideally, synthesizing an image should maximize the distances in response space for all pairs of ensemble DNNs, but this is computationally time-consuming. Instead, we approximately maximize the ensemble disagreement by synthesizing an image in the following procedure. We intialize a candidate image with a randomly-chosen normal image. Then, for each of 20~iterations, we randomly choose two ensemble DNNs, compute the Euclidean distance between the predicted response vectors of the two ensemble DNNs, and use this distance as an objective function for which we can optimize by performing backprop through the network to update the pixel intensities (i.e., the weights of the ensemble DNNs remain fixed). However, because updating pixel intensities directly leads to adversarial examples \cite{goodfellow2014}, we further backprop through a generator network that maps a feature vector to an image. The idea is to synthesize images by taking a gradient step along the manifold of natural images (i.e., feature space) rather than a gradient step in pixel space. We use a similar network architecture and GAN training procedure from a previous study for the generator and discriminator networks \cite{dosovitskiy2016}. As feature vectors, we use the output features of the ResNet50 sub-network, and we use the same sub-network for the comparator network. We train the generator and discriminator networks on our 12~million image dataset. Because we perform batch active learning, we also incorporate a diversity term in the objective function equal to the minimum Euclidean distance between the feature vector of the current synthesized image and all previously-converged synthesized feature vectors for that session. This term encourages diversity by ensuring that the current synthesized feature vector is far from any previously-synthesized feature vectors. To recover the 250~synthesized images, we pass these feature vectors through the generator network. 
        
    \end{itemize}

\clearpage


\renewcommand{\figurename}{\textbf{Supplementary Figure}}
\setcounter{figure}{0}

\section*{Appendix}

Here we formally derive the active learning theory in Equation~\ref{eqn:new_point}. This theory presents the optimal strategy for training a linear regression with active learning (assuming the ground truth mapping is also linear). A counterintuitive result from this theory is that the optimal stimuli can be chosen \emph{before} collecting training data as opposed to typical active learning algorithms that optimize stimuli \emph{during} the data collection and training process. We have adapted this theory from previous work \cite{pukelsheim2006, jaakkola2006} and present it in the context of predicting neural responses from images. 

Let us assume a linear mapping between image $\x \in \mathbb{R}^K$ (for $K$ pixels) and neural response $y$: $y = \beta^T \x + \epsilon$, with weight vector $\beta \in \mathbb{R}^K$ and noise $\epsilon \sim \mathcal{N}(0,\sigma_\epsilon^2)$. We model this mapping with $\hat{y} = \hat{\beta}^T \x$, where $\hat{\beta} = \inv{\Sigma} \X \y$ for a set of (re-centered) training images $\X \in \mathcal{R}^{K \times N}$ (for $K$ pixels and $N$ images), unnormalized covariance matrix $\Sigma = \X \X^T$, and responses $\y \in \mathbb{R}^N$. We can compute the expected error that represents how well our estimate $\hat{\beta}$ matches that of ground truth $\beta$:

\begin{equation}
\begin{split}
    \textrm{E}\Big[\big\|\beta - \hat{\beta}\big\|_2^2 \mid \X \Big] = 
    \textrm{E}\Big[\big\|\beta - \inv{\Sigma} \X \y \big\|_2^2 \Big] = 
    \textrm{E}\Big[\big\|\beta - \inv{\Sigma} \X \big(\X^T\beta + 
    \underline{\epsilon}\big) \big\|_2^2 \Big] = \\
    \textrm{E}\Big[\big\| \inv{\Sigma} \X \underline{\epsilon} \big\|_2^2 \Big] =
    \textrm{E}\Big[\textrm{Tr}\big[(\inv{\Sigma} \X \underline{\epsilon})^T (\inv{\Sigma} \X \underline{\epsilon})\big]\Big] = 
    \sigma_\epsilon^2 \textrm{ Tr}(\inv{\Sigma})
\end{split}
\label{eqn:error}
\end{equation}

where $\underline{\epsilon} \in \mathbb{R}^N$ is the vector of noise values for each response. We seek a new, unshown image $\x_\textrm{next}$, from the set of all possible (re-centered) images $\mathcal{X}$, that minimizes the error in Eqn.~\ref{eqn:error} when we train $\hat{\beta}$ on the new image $\x_\textrm{next}$ and its response $y$. Importantly, we do not know $y$, so we cannot ask which image has the largest prediction error $\|y - \hat{y}(\x_\textrm{next})\|_2^2$. Instead, we choose the image $\x_\textrm{next}$ that minimizes the error between the true $\beta$ and our estimate $\hat{\beta}$ trained on images $[\X, \x_\textrm{next}] \in \mathbb{R}^{K \times (N+1)}$:

\begin{equation}
\begin{split}
    \x_\textrm{next} =
    \min\limits_{\x \in \mathcal{X}} \textrm{E}\Big[\big\|\beta - \hat{\beta}\big\|_2^2 \mid \big[\X, \x\big] \Big] =
    \min\limits_{\x} \sigma_\epsilon^2 \textrm{ Tr}\big[(\Sigma + \x\x^T)^{-1} \big] \overset{(\triangle)}{=} \\
    \min\limits_{\x} \textrm{Tr}(\inv{\Sigma}) - \frac{\x^T \inv{\Sigma} \inv{\Sigma} \x}{1 + \x^T \inv{\Sigma} \x} =
    \max\limits_{\x} \frac{\x^T \inv{\Sigma} \inv{\Sigma} \x}{1 + \x^T \inv{\Sigma} \x}
\end{split}
\label{eqn:full_deriv}
\end{equation}

where in ($\triangle$) we apply the Sherman-Morrison formula. The objective function of the rightmost formula in Eqn.~\ref{eqn:full_deriv} reveals that the image $\x$ that decreases the error the most is the one that maximizes the magnitude of its projection along the eigenvectors of $\inv{\Sigma}$ with the largest eigenvalues. An easier way to intuit this optimization is to assume that the $K$ pixels of $\x$ are uncorrelated (e.g., via a change of basis). Under this setting, the covariance matrix of pixel intensities $\Sigma$ is a diagonal matrix with on-diagonal entries $\Sigma_{k,k} = \sigma_k^2$. The optimization becomes the following:

\begin{equation}
    \x_{\textrm{next}} = 
    \argmax\limits_{\x \in \mathcal{X}} \frac{\x^T (\inv{\Sigma})^2 \x}{1 + \x^T \inv{\Sigma} \x} \; \overset{\x_i \perp \x_j}{=} \;
    \argmax\limits_{\x} \sum\limits_{k=1}^{K} \nicefrac{\x_k^2}{\sigma_k^2}
\end{equation}

where $\x_i \perp \x_j$ indicate the $i$th and $j$th variables in $\x$ are uncorrelated for all $i$ and $j$, $i \neq j$. Intuitively, we seek the image $\x$ that maximizes the magnitude of its projection along dimensions in pixel space with small variance $\sigma_\textrm{small}^2$. We do this in order to increase the strength of the weakest signal $\sigma_\textrm{small}^2$ relative to noise $\sigma_\epsilon^2$ (i.e., increase the signal-to-noise ratio $\sigma_\textrm{small}^2 / \sigma_\epsilon^2$). Note that the optimization in Eqn.~\ref{eqn:full_deriv} does not depend on previous responses $\y$ nor the current model's weights $\hat{\beta}$. Thus, $\x_\textrm{next}$ can be chosen \emph{before} training the model.

\clearpage

\begin{figure}
    \includegraphics[width=\textwidth]{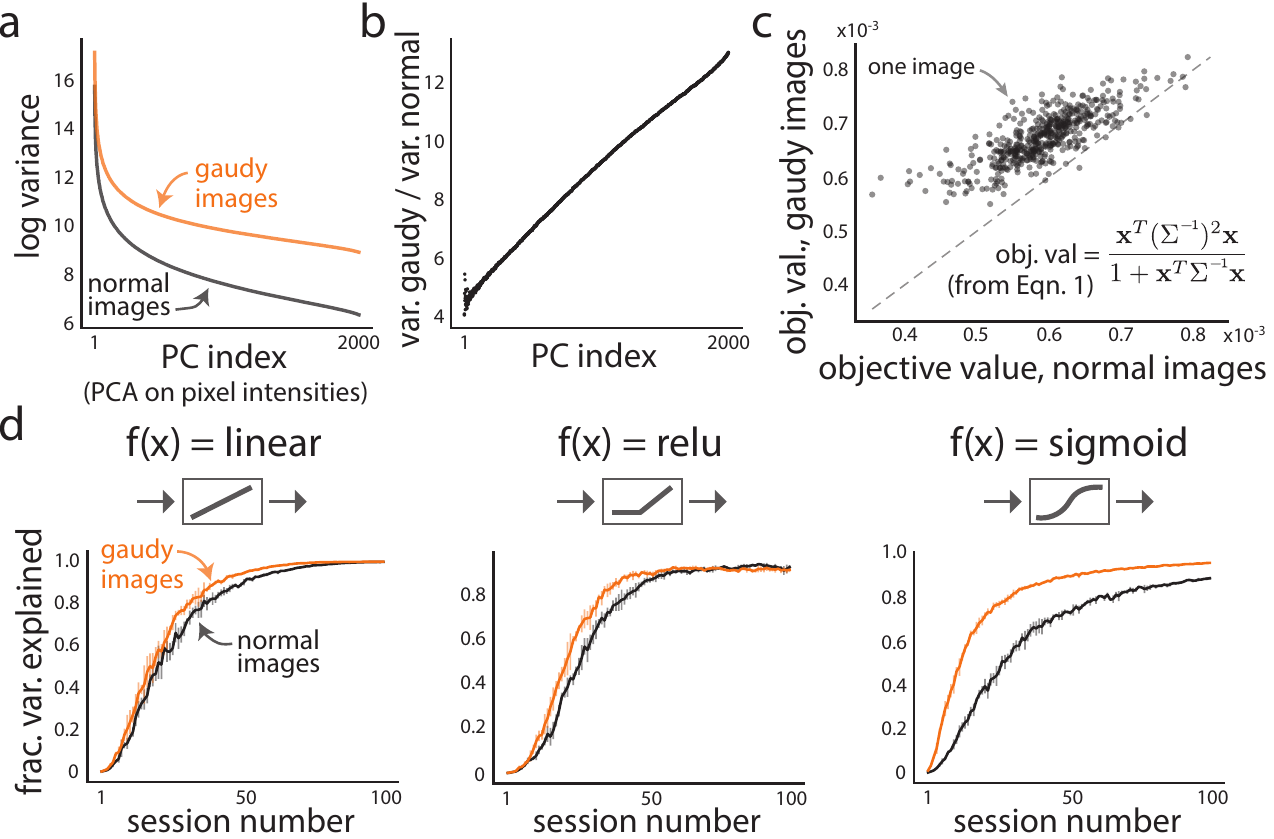}
    \caption{
    Assessing the extent to which gaudy images increase the variance of pixel dimensions and whether performance plateaus after training on gaudy images for many sessions.\\
    \a. We apply PCA to 10,000 grayscale images ($112 \times 112$ pixels) and compute the variance (i.e., eigenvalue) for each PC dimension, up to the top 2,000 PCs. Gaudy images (orange line) yield larger variances than those of normal images (black line). PCA is applied separately to either normal or gaudy images.\\
    \b. Ratios of variance explained (variance of gaudy images divided by variance of normal images) for each PC index. Gaudy images yield larger variances for pixel dimensions that explain a small amount of variance than those of normal images (i.e., ratios increase with PC index). This suggests that gaudy images satisfy the optimal strategy derived from the active learning theory in Eqn.~\ref{eqn:new_point}: Maximize the variance for low-variance pixel dimensions.\\
    \c. We further confirm that gaudy images yield larger objective values of Eqn.~\ref{eqn:new_point} than those of normal images (dots are above gray dashed line, $p < 0.002$, permutation test). In the objective function, $x$ is a (re-centered) image with $K$ pixels, and $\Sigma$ is the $K \times K$ covariance matrix of pixel intensities.\\
    \d. Same as in Fig.~\ref{fig:sec3_glm_results}\b-\d\ except that we train for 100~sessions instead of 30~sessions. For the linear (left panel) and relu (middle panel) activation functions, performance plateaus after \texttildelow50~sessions with no appreciable difference in the final performance between training either on normal or gaudy images. For the sigmoid activation function (right panel), more sessions are needed to reach a plateau. Error bars indicate 1~s.d. for 5~runs (some error bars are too small to see).
    }
    \label{fig:suppfig1_glm_results}
\end{figure}

\clearpage

\begin{figure}
    \centering
    \includegraphics[width=\textwidth]{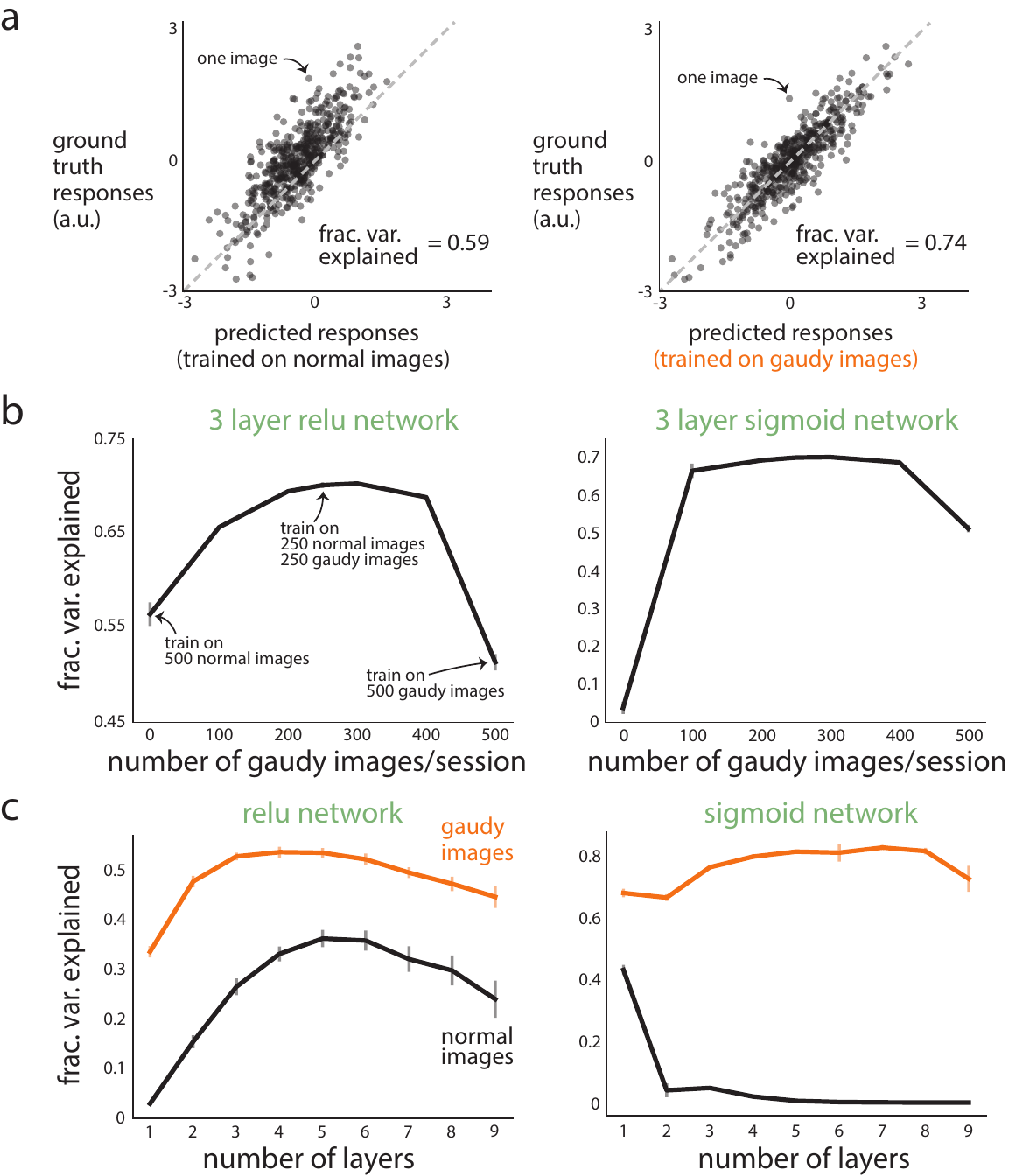}
    \caption{Additional results for using gaudy images to train DNNs: Assessing the residual error of predictions, changing the number of gaudy images trained per session, and varying the number of DNN layers. (continued on next page...)}
    \label{fig:suppfig2_dnn_results_sweep}
\end{figure}

    \begin{figure}
        \contcaption{
    (...continued from previous page.) \\
    \a. Predicted responses versus ground truth responses for an example simulated neuron from VGG19 (same neuron in both panels). Training on gaudy images (right panel) leads to smaller errors between predicted and ground truth responses than those for training on normal images (left panel). This is especially true for outliers (i.e., ground truth responses with large magnitudes). Predicted responses are from training a 3~layer convolutional network with relu activation functions (same readout network as in Fig.~\ref{fig:sec4_dnn_results}\c) after 30~sessions. Each dot corresponds to one heldout normal image. Fraction variance explained (`frac. var. explained') is computed from responses to 4,000~heldout normal images. This example neuron has the median fraction variance explained of all VGG19 surrogate neurons when trained on normal images.  \\
    \b. For each session, we train a DNN with 500~images. One design choice is the number of gaudy images to use per session (the rest being normal images). We test this by training (for 30~sessions) a 3~layer convolutional network either with relu activation functions (left panel) or with sigmoid activation functions (right panel) to predict VGG19 responses to heldout normal images. We find that the setting in which all 500~training images are gaudy images does not achieve the best performance (rightmost point in each panel). Instead, a mix between gaudy and normal images achieves the best performance (e.g., 250~normal and 250~gaudy images). The reason this mix performs best is likely because responses to gaudy images tend to be outliers (Fig.~\ref{fig:sec5_why_gaudi}\a), and training on only gaudy images leads to a large mismatch between the training and test set distributions. Instead, a 50\%-50\%\ mix between gaudy and normal images represents a trade-off between training on samples with large errors (i.e., gaudy images) versus ensuring that the training set distribution matches the test set distribution (i.e., normal images). Thus, this mix is likely an important ingredient to train a DNN with data augmentation or active learning methods. We use this 50\%\-50\%\ mix to train readout networks in all other analyses in this paper.\\
    \c. We vary the number of layers of the readout network (i.e., a $K$-layer convolutional network iwth 512~filters per layer, similar to readout network in \b) from 1 to 9~layers. We find that training on gaudy images outperforms training on normal images for all numbers of layers (orange line above black line). This includes networks with relu (left panel) and sigmoid (right panel) activation functions. Networks are trained to predict VGG19 responses. We use a learning rate of 1e-1 for all numbers of layers for the relu networks (trained for 20~sessions) and a learning rate of 5e-1 for the sigmoid networks. We find that the deeper sigmoid networks (e.g., 6 or more layers) require more training data, so we increase the number of sessions to equal the number of layers multiplied by 10 (e.g., for 10~layers, we train for 100~sessions), except for 1~layer (for which we trained for 20~sessions). \\
    For panels \b\ and \c, error bars indicate 1~s.d. for 5~runs (some error bars are too small to see).
        }
    \end{figure}

\clearpage

\begin{figure}
    \centering
    \includegraphics[width=\textwidth]{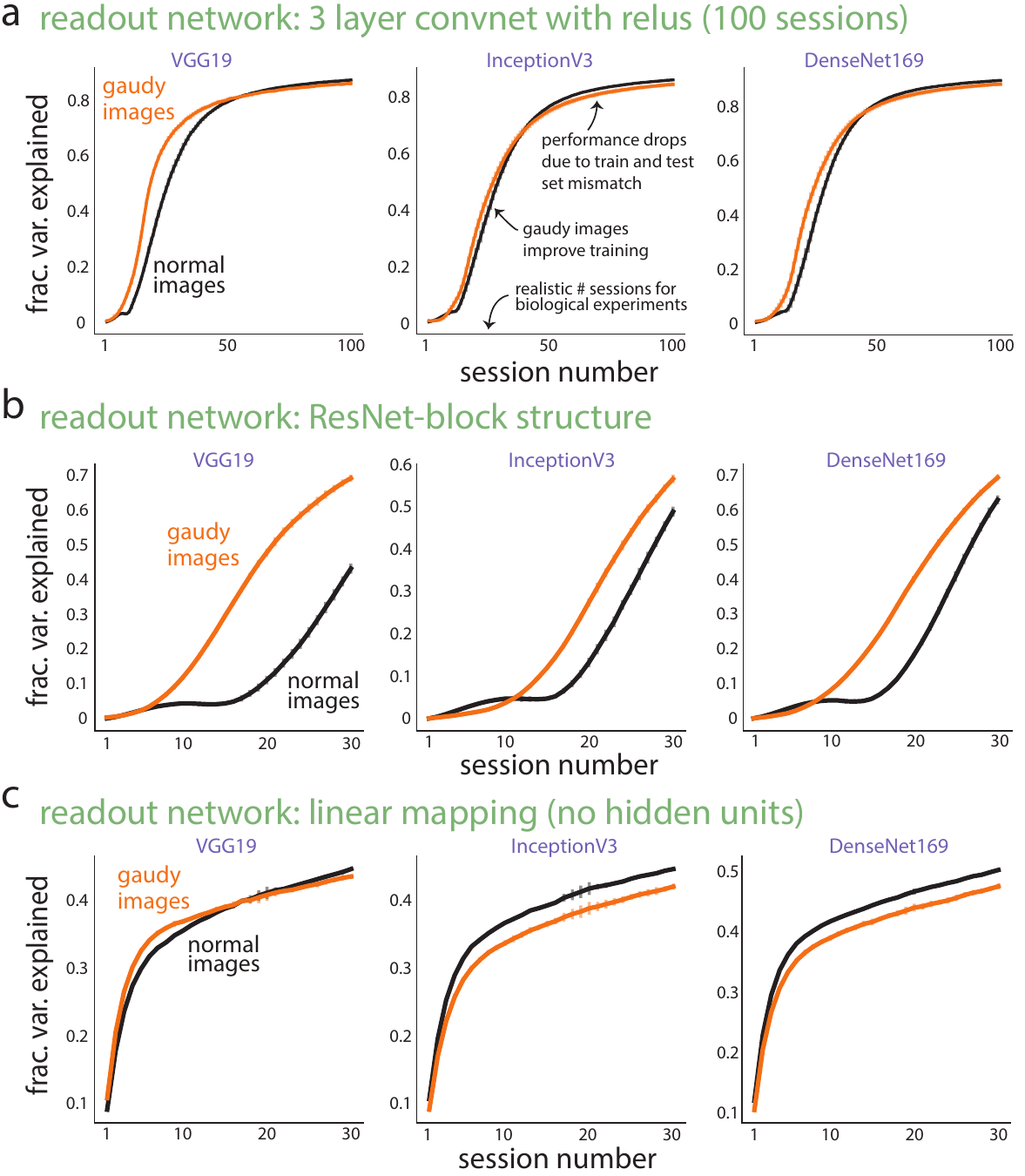}
    \caption{Additional results for using gaudy images to train DNNs: Training a DNN for many sessions, training a DNN with a ResNet-like architecture, and training a linear mapping. (continued on next page...)}
    \label{fig:suppfig3_dnn_results_fracvars}
\end{figure}

    \begin{figure}
        \contcaption{
        (...continued from previous page.) \\
    \a. Results for training the same readout network as in Fig.~\ref{fig:sec4_dnn_results}\c\ 
        (i.e., 3~layer convolutional network with relu activation functions) for 100~sessions. Performances for the first 30~sessions are the same as those in Fig.~\ref{fig:sec4_dnn_results}\c. However, after \texttildelow50~sessions, performance plateaus, and we observe a slight increase in performance for training on normal images versus training on gaudy images (black line slightly above orange line after 50~sessions). This is expected, as gaudy images represent a mismatch between training and test set distributions, whereas no mismatch exists when training on a large number of normal images. Thus, adopting a hybrid approach of initially training on gaudy images and then transitioning to training on normal images will likely yield the best results. For training DNNs to predict visual cortical responses, we suspect the number of training images needed to begin to transition to training only on normal images (here, 25,000 images) is likely out of reach for most neuroscientific experiments.\\
    \b. Results for a readout network with 4~ResNet blocks \citep{he2016}. We find that gaudy 
        images (orange) substantially improve performance over normal images (black) for this network architecture. Each ResNet block consists of 3~convolutional layers, where the first two layers have 256~filters ($3 \times 3$ kernel, stride of 1, separable convolution) and the last layer has 512~filters ($1 \times 1$ kernel, stride of 1, separable convolution). We include a skip connection for the last operation of each block, which sums the output of the block's last layer to the block's input. The first layer of the network is a convolutional layer with 512~filters ($1 \times 1$ kernel, linear activation function), whose output is then passed to the first ResNet block. The final layer is a linear mapping between the output of the last ResNet block and the predicted responses. We initialize the network randomly with default Keras settings, including the sampling of initial weights from the Glorot uniform distribution. We use a learning rate of 2e-1 and a momentum value of 0.7. The testing procedure is the same as that for Figure~\ref{fig:sec4_dnn_results}\c\ and \d. \\
    \c. For completeness, we also train a linear mapping for the readout network.
        Here, because the true mapping between features and responses is likely nonlinear, we do \emph{not} expect gaudy images to improve training. This is because the active learning theory for linear regression (Eqn.~\ref{eqn:new_point}) assumes that the ground truth mapping between features and responses is linear---which is not the case here. We fit the linear mapping using SGD with a learning rate of 1e-2 and a momentum value of 0.7. \\
        In our work, we use a nonlinear mapping (i.e., the readout network in \a) between the features of a pre-trained DNN and responses. However, a linear mapping may perform as well as or better than a nonlinear mapping. Indeed, for a small amount of training data, we find this to be the case (compare performance for VGG19, session 10, between \a\ and \c, left panels). However, when we train both mappings for 100~sessions, the performance of the readout network (fraction variance explained of $0.873 \pm 0.002$) is larger than that of a linear mapping ($0.595 \pm 0.001$, where $\pm$ indicates 1~s.d.). Thus, a nonlinear mapping, trained on enough data, is more appropriate than a linear mapping for models of visual cortical neurons. \\
        For all panels, error bars indicate 1~s.d. for 5~runs (some error bars are too small to see).
        }
    \end{figure}

\clearpage

\begin{figure}
    \centering
    \includegraphics[width=5in]{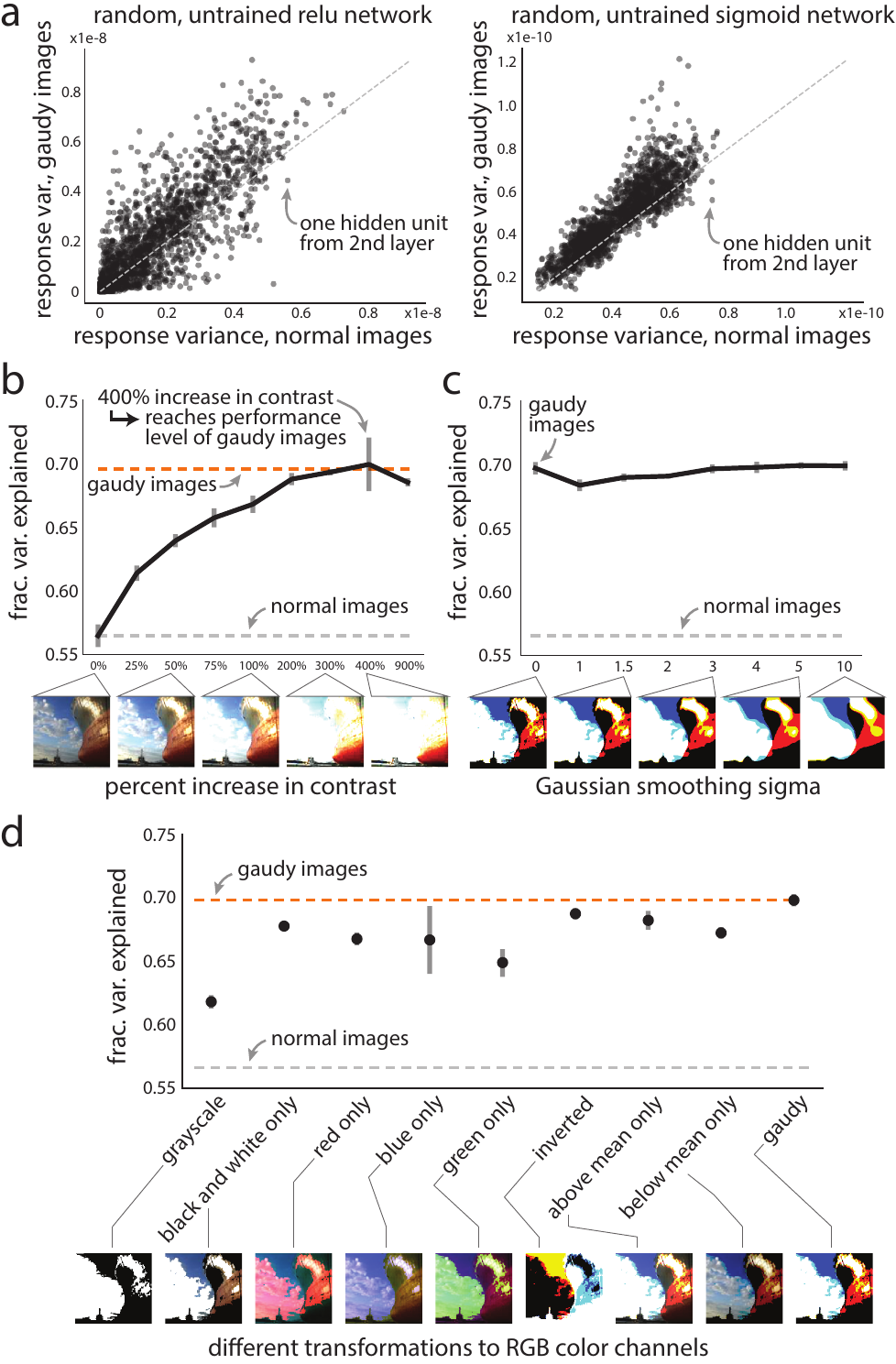}
    \caption{
    Additional results to understand which features of gaudy images are the most important for improving training.\\
    \a. To understand how gaudy images affect optimization of DNN weights, we consider 
        to what extent gaudy images drive diverse activity in the readout network. Diverse activity is advantageous for optimization, as the input variables of a hidden unit need to vary in order to regress to the desired output variable. 
    (continued on next page...)}
    \label{fig:suppfig4_why_gaudy}
\end{figure}

\clearpage

\textbf{Supplementary Figure~\ref{fig:suppfig4_why_gaudy}:} 
(...continued from previous page.) \\
Here, we compute the variance of responses to either 1,000~normal or 1,000~gaudy images in  randomly-initialized, untrained readout networks (same architectures as in Fig.~\ref{fig:sec4_dnn_results}\c\ and \d). We find that gaudy images yield significantly larger output variances for hidden units in the second layer, both for a randomly-initialized relu network (left panel, $p<0.001$, permutation test) and a randomly-initialized sigmoid network (right panel, $p<0.001$, permutation test). This suggests that gaudy images lead to more variation in the input variables of each hidden unit (especially for the small number of images in a batch), which in turn makes it easier to identify the most informative input variables for that hidden neuron. \\
    \b. The gaudy transformation is akin to substantially increasing the contrast of a normal image. 
        Here, we train the readout network on images with varying levels of contrast to infer the extent to which the gaudy transformation increases contrast. We change contrast with the commonly-used contrast stretching approach \citep{davies2004}. Consider a percent contrast increase $c$ (e.g., $c=10$ indicates a 10\%\ increase in contrast). For each image, we compute a minimum luminance $m$ as the 5\%\ quantile of pixel intensities for that image. Then, for each pixel intensity $p$, we compute a new pixel intensity $\tilde{p} = (p-m)(100 + c)/100$, and then clip $\tilde{p}$ to be between 0 and 255. The bottom inset is a set of example images for different percent increases in contrast. The training and testing procedure, as well as the readout network, are the same as in Figure~\ref{fig:sec5_why_gaudi}\d, \e, and \f. We find that gaudy images represent an increase of 400\%\ in contrast, as 400\%\ achieves similar performance as that of gaudy images. This increase is substantially larger than that used in data augmentation methods, typically no more than 50\%\ \cite{cubuk2019}. \\
    \c. In Figure~\ref{fig:sec5_why_gaudi}\d, we transform an image to a gaudy image and 
        then perform Gaussian smoothing, eliminating any high-contrast edges. Here, we perform this procedure in reverse (i.e., first smooth a normal image and then transform to gaudy) to further test if the high-contrast edges of gaudy images are important features to improve training. A Gaussian smoothing sigma of 1.0 corresponds to a s.d. of 1~pixel. Under this procedure, the smoothed gaudy images still retain high-contrast edges but lose high-frequency spatial information such as textures (inset, rightmost example image). Even for highly-smoothed gaudy images, performance is similar to that of gaudy images (cf. sigma values of 0 and 10). This results provides further evidence that high-contrast edges are likely the most important features of gaudy images to improve training. \\
    \d. We also test if different color transformations lead to different training
        performances (bottom inset is a set of example images, one for each transformation). We find that these color transformations do improve performance above that of normal images (gray dashed line) but not above that of gaudy images (orange dashed line and rightmost dot). 
        The image transformations are as follows (with $\bar{p}$ as the mean pixel intensity of the image):
        \begin{itemize}
            \item \textbf{grayscale}: Transform each pixel intensity to its across-channel mean intensity, then transform to gaudy (see \textbf{gaudy} below).
            \item \textbf{black and white only}: Transform to gaudy only pixels whose intensities across channels are either all above $\bar{p}$ or all below $\bar{p}$. Pixels that do not satisfy these criteria are not transformed and remain normal.
            \item \textbf{red only}: Transform to gaudy the pixel intensities of the red-channel only (green and blue channel intensities remain normal) with threshold $\bar{p}$.
            \item \textbf{green only}: Same transformation as \textbf{red only} except only for pixel intensities of the green-channel.
            \item \textbf{blue only}: Same transformation as \textbf{red only} except only for pixel intensities of the blue-channel.
            \item \textbf{inverted}: An inversion of the gaudy transformation. If pixel intensity $p < \bar{p}$, the gaudy pixel intensity $\tilde{p} = 255$, else $\tilde{p} = 0$. 
            \item \textbf{above mean only}: If pixel intensity $p \geq \bar{p}$, then the gaudy pixel intensity $\tilde{p} = 255$, else $\tilde{p} = p$. The resulting image emphasizes ``bright'' regions of the image.
            \item \textbf{below mean only}: Opposite of \textbf{above mean only}. If $p < \bar{p}$, then $\tilde{p} = 0$, else $\tilde{p} = p$. The resulting image emphasizes ``dark'' regions of the image.
            \item \textbf{gaudy}: The gaudy transformation. If pixel intensity $p < \bar{p}$, then we set the gaudy pixel intensity $\tilde{p}$ to $\tilde{p} = 0$, else $\tilde{p} = 255$.
        \end{itemize}
        For panels \b, \c, and \d, error bars indicate 1~s.d. for 5~runs (some error bars are too small to see).

\clearpage

\begin{figure}
    \centering
    \includegraphics[width=5in]{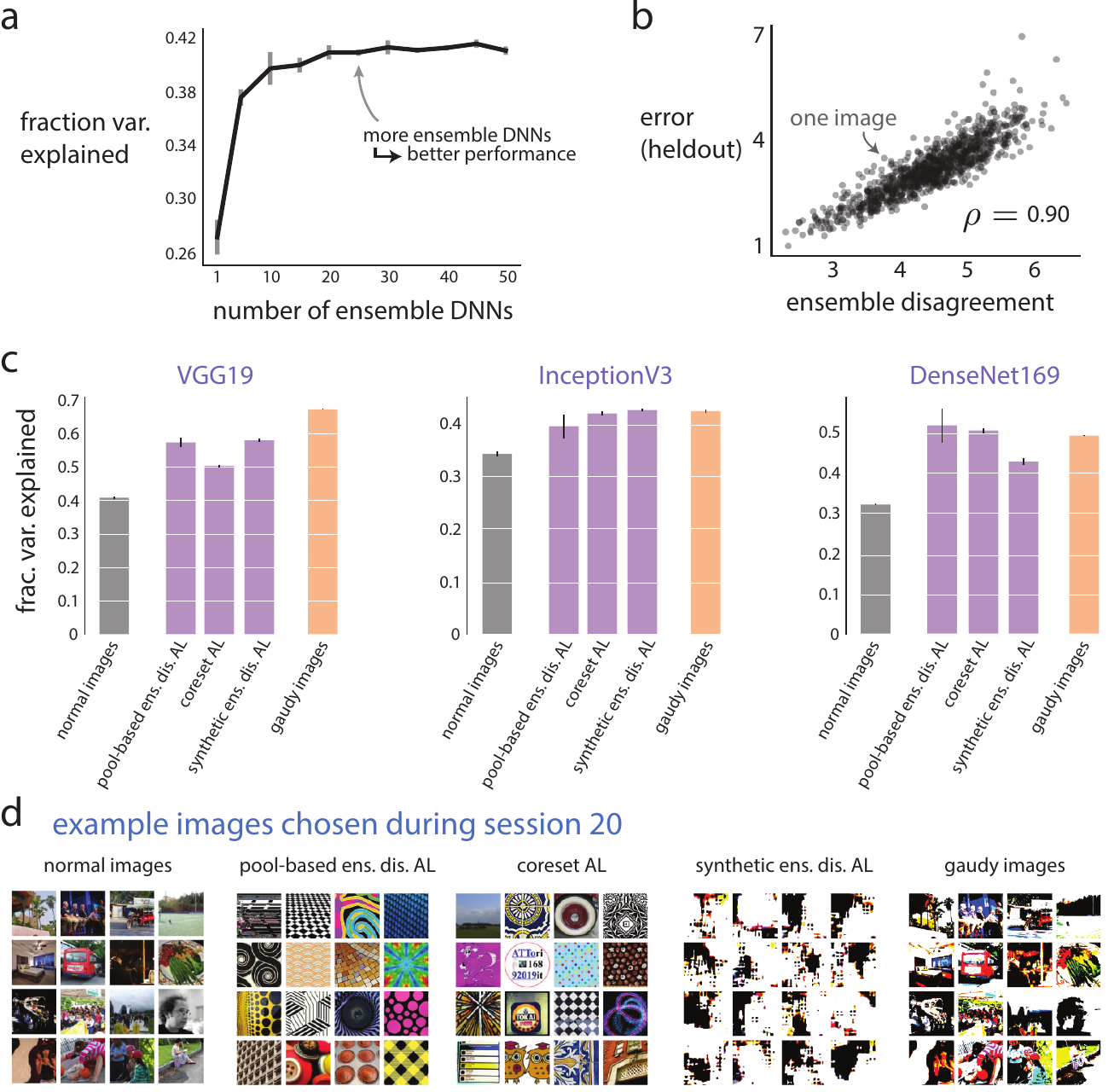}
    \caption{
    Gaudy images, chosen \emph{before} training, improve the training of DNNs on par with or greater than active learning (AL) algorithms, which choose images \emph{during} training. \\
    (continued on next page...)}
    \label{fig:suppfig5_AL_ensemble_results}
\end{figure}

    \begin{figure}[t!]
        \contcaption{
        (...continued from previous page.) \\
        \a. For the active learning experiments, we use an ensemble of readout DNNs 
            (Fig.~\ref{fig:sec6_AL_results}\a), primarily because we can then use the disagreement among the ensemble DNNs as a measure of uncertainty for active learning. Here, we vary the number of ensemble DNNs and train them (over 20~sessions with normal images, where each ensemble DNN is a 3~layer convolutional network with relu activation functions) to predict ground truth VGG19 responses. We find that increasing the number of ensemble DNNs improves performance (compare 1 to 25~ensemble DNNs). We use 25~ensemble DNNs in all other analyses in Figure~\ref{fig:sec6_AL_results} and this figure. \\
        \b. The ensemble disagreement (defined as the median distance of predicted
            response vectors among ensemble DNNs, see Methods) is significantly correlated to the error of heldout images (Pearson's correlation $\rho=0.90$, $p < 0.001$, permutation test). Thus, choosing images with high ensemble disagreement (for which we have access during training) is akin to choosing images with large error (for which we do \emph{not} have access during training). Images with large error yield larger, more informative gradient steps which in turn train the networks with less training data. \\
        \c. Performance results for a large range of AL algorithms across different
            pre-trained DNNs used to simulate visual cortical responses. The AL algorithms (purple) are the same as in Fig.~\ref{fig:sec6_AL_results}\b. Models are trained in the same manner as in Figure~\ref{fig:sec6_AL_results}\b\ for 20~sessions. We find that gaudy images either outperform or are on par with the tested active learning algorithms (orange bars above or equal to other bars). This suggests that gaudy images emphasize natural image features (e.g., high contrast edges) that aid in training, and that these features are the most important for training (else AL algorithms would have identified other features to increase their performance over that of gaudy images). \\
        Error bars in \a\ and \c\ indicate 1~s.d. over 5~runs (some error bars are too small to see). \\
        \d. Example images chosen or synthesized by the different AL algorithms 
            on session~20 when predicting VGG19 responses. The pool-based ens. dis. AL and coreset AL algorithms tend to find bright, textured images, while the synthetic ens. dis. AL algorithm appears to synthesize gaudy-like images.
        }
    \end{figure} 

\clearpage

\begin{figure}
    \centering
    \includegraphics[width=4.75in]{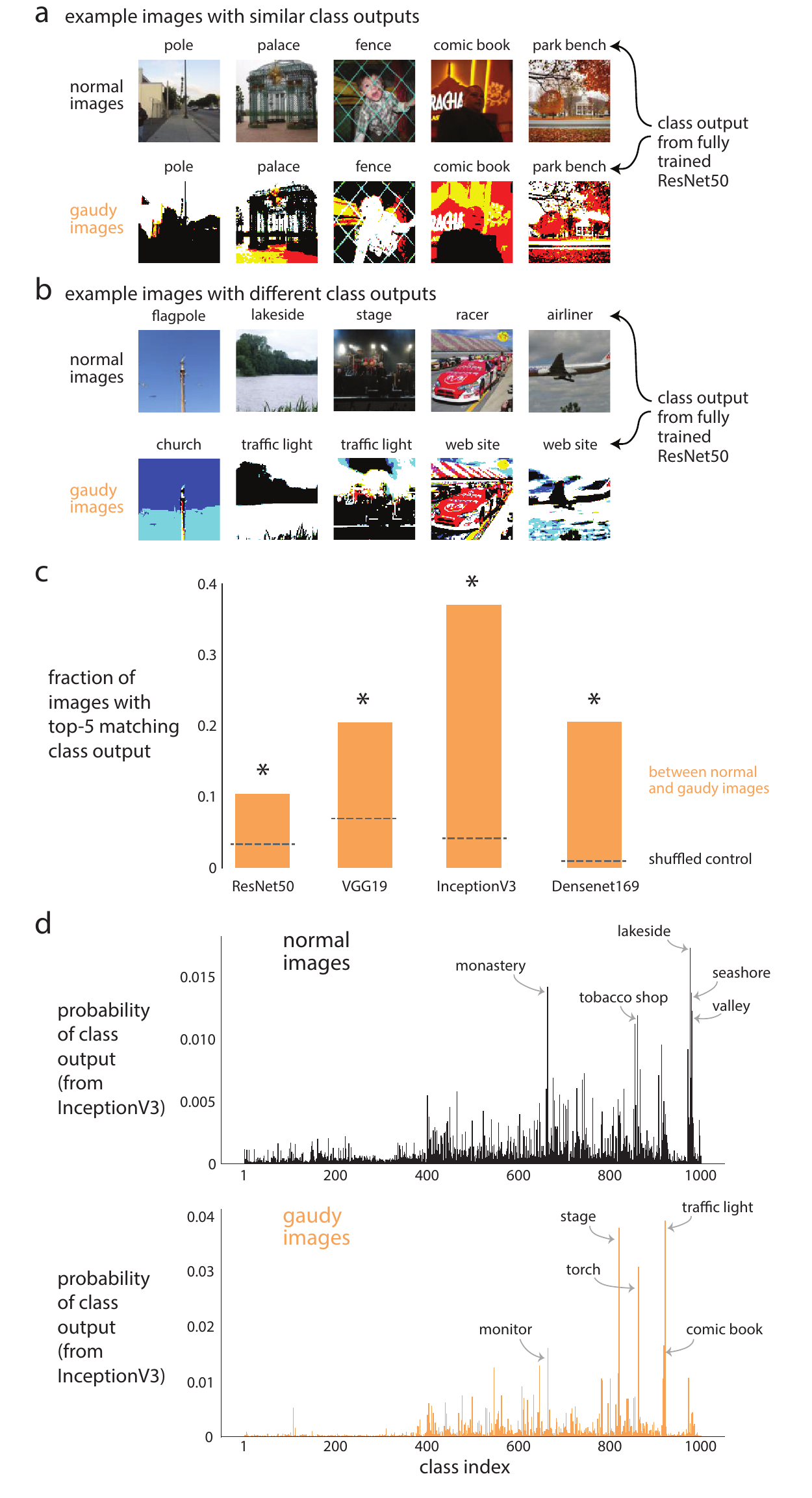}
    \caption{(Caption on next page.)}
    \label{fig:suppfig6_classification_results}
\end{figure}

    \begin{figure}[t!]
        \contcaption{
        (Figure on previous page.) \\
        Gaudy images may be helpful for data augmentation (especially if used in tandem with transfer learning). Data augmentation assumes that for a given image, the output class label is invariant to different image transformations (such as rotations or translations). Here, we ask if gaudy images may be useful for data augmentation. To test this, we ideally would have humans label normal images and their corresponding gaudy images to see if the human labels are invariant to the gaudy transformation. Because that process is costly, as a first step we test whether a fully-trained DNN is likely to output the same class label for a normal image as for its corresponding gaudy image. \\
        \a. We pass example normal images and their corresponding gaudy images 
            through ResNet-50, fully-trained on ImageNet. We find example images whose top-5 classes (i.e., the top 5~classes with the largest prediction probabilities for each image) have at least one matching label between the normal and gaudy versions of the same image.\\
        \b. Same as in \a, except that the example images have no matching top-5 classes.
            ResNet-50 misclassified the gaudy images probably due to their bright colors (e.g., `traffic light' and `web site' invoke brightly-colored images), although a human annotator would likely classify these images correctly. \\
        \c. We compute the fraction of images with at least one matching class 
            between the top-5 predicted classes of a normal image and a gaudy image. Classes are predicted from one of four different DNNs pre-trained for object recognition (orange, fraction estimated with 5,000~images). See Methods for details about the pre-trained DNNs. These fractions are significantly greater ($p < 0.001$, proportion of runs) than those of a shuffled control (gray dashed lines, means across runs), where for each run we shuffle the predicted labels of the gaudy images and re-compute the fraction of matching top-5 classes. This result suggests that including gaudy images for data augmentation may improve training, as many gaudy images still retain enough information to be correctly classified. \\
        \d. One potential confound in the analysis in \c\ is that the 5,000~images 
            come from the FlickR image dataset and are not grouped into classes with equal numbers of instances. Thus, it could be the case that these images have a class imbalance (e.g., more images contain animate objects than images that contain inanimate objects), which would bias our estimates in \c. To control for this, we sum the predicted class probabilities for one pre-trained DNN (InceptionV3) over all 5,000~images for the normal images (top panel) and gaudy images (bottom panel) and normalize these sums by dividing by the number of images. We find that the class predictions of normal images are somewhat spread evenly across classes (top panel, black bars), suggesting the FlickR image dataset does not have a strong class imbalance. The class predictions of gaudy images are more confined to a group of classes (bottom panel, classes with indices between 400~and~1000 have most of the probability). In addition, the classes with the largest probabilities for normal images are all different from those for gaudy images (class labels with arrows pointing to corresponding bars). These results indicate gaudy images are more prone to be misclassified than normal images for different groups of classes (e.g., animate versus inanimate objects). Overall, the results from this figure suggest gaudy images are likely to be useful for data augmentation when training DNNs for object recognition and transfer learning tasks (where training data is limited).
        }
    \end{figure} 


\begin{thebibliography}{65}
\providecommand{\natexlab}[1]{#1}
\providecommand{\url}[1]{\texttt{#1}}
\expandafter\ifx\csname urlstyle\endcsname\relax
  \providecommand{\doi}[1]{doi: #1}\else
  \providecommand{\doi}{doi: \begingroup \urlstyle{rm}\Url}\fi

\bibitem[Heeger et~al.(1996)Heeger, Simoncelli, and Movshon]{heeger1996}
David~J Heeger, Eero~P Simoncelli, and J~Anthony Movshon.
\newblock Computational models of cortical visual processing.
\newblock \emph{Proceedings of the National Academy of Sciences}, 93\penalty0
  (2):\penalty0 623--627, 1996.

\bibitem[DiCarlo et~al.(2012)DiCarlo, Zoccolan, and Rust]{dicarlo2012}
James~J DiCarlo, Davide Zoccolan, and Nicole~C Rust.
\newblock How does the brain solve visual object recognition?
\newblock \emph{Neuron}, 73\penalty0 (3):\penalty0 415--434, 2012.

\bibitem[Yamins et~al.(2014)Yamins, Hong, Cadieu, Solomon, Seibert, and
  DiCarlo]{yamins2014}
Daniel~LK Yamins, Ha~Hong, Charles~F Cadieu, Ethan~A Solomon, Darren Seibert,
  and James~J DiCarlo.
\newblock Performance-optimized hierarchical models predict neural responses in
  higher visual cortex.
\newblock \emph{Proceedings of the National Academy of Sciences}, 111\penalty0
  (23):\penalty0 8619--8624, 2014.

\bibitem[Klindt et~al.(2017)Klindt, Ecker, Euler, and Bethge]{klindt2017}
David Klindt, Alexander~S Ecker, Thomas Euler, and Matthias Bethge.
\newblock Neural system identification for large populations separating
  “what” and “where”.
\newblock In \emph{Advances in Neural Information Processing Systems}, pages
  3506--3516, 2017.

\bibitem[Abbasi-Asl et~al.(2018)Abbasi-Asl, Chen, Bloniarz, Oliver, Willmore,
  Gallant, and Yu]{abbasi2018}
Reza Abbasi-Asl, Yuansi Chen, Adam Bloniarz, Michael Oliver, Ben~DB Willmore,
  Jack~L Gallant, and Bin Yu.
\newblock The deeptune framework for modeling and characterizing neurons in
  visual cortex area v4.
\newblock \emph{bioRxiv}, page 465534, 2018.

\bibitem[Sinz et~al.(2018)Sinz, Ecker, Fahey, Walker, Cobos, Froudarakis,
  Yatsenko, Pitkow, Reimer, and Tolias]{sinz2018}
Fabian Sinz, Alexander~S Ecker, Paul Fahey, Edgar Walker, Erick Cobos,
  Emmanouil Froudarakis, Dimitri Yatsenko, Zachary Pitkow, Jacob Reimer, and
  Andreas Tolias.
\newblock Stimulus domain transfer in recurrent models for large scale cortical
  population prediction on video.
\newblock In \emph{Advances in Neural Information Processing Systems}, pages
  7199--7210, 2018.

\bibitem[Schrimpf et~al.(2018)Schrimpf, Kubilius, Hong, Majaj, Rajalingham,
  Issa, Kar, Bashivan, Prescott-Roy, Schmidt, et~al.]{schrimpf2018}
Martin Schrimpf, Jonas Kubilius, Ha~Hong, Najib~J Majaj, Rishi Rajalingham,
  Elias~B Issa, Kohitij Kar, Pouya Bashivan, Jonathan Prescott-Roy, Kailyn
  Schmidt, et~al.
\newblock Brain-score: Which artificial neural network for object recognition
  is most brain-like?
\newblock \emph{BioRxiv}, page 407007, 2018.

\bibitem[Cadena et~al.(2019)Cadena, Denfield, Walker, Gatys, Tolias, Bethge,
  and Ecker]{cadena2019}
Santiago~A Cadena, George~H Denfield, Edgar~Y Walker, Leon~A Gatys, Andreas~S
  Tolias, Matthias Bethge, and Alexander~S Ecker.
\newblock Deep convolutional models improve predictions of macaque v1 responses
  to natural images.
\newblock \emph{PLoS computational biology}, 15\penalty0 (4):\penalty0
  e1006897, 2019.

\bibitem[Kindel et~al.(2019)Kindel, Christensen, and Zylberberg]{kindel2019}
William~F Kindel, Elijah~D Christensen, and Joel Zylberberg.
\newblock Using deep learning to probe the neural code for images in primary
  visual cortex.
\newblock \emph{Journal of vision}, 19\penalty0 (4):\penalty0 29--29, 2019.

\bibitem[Zhang et~al.(2019)Zhang, Lee, Li, Liu, and Tang]{zhang2019}
Yimeng Zhang, Tai~Sing Lee, Ming Li, Fang Liu, and Shiming Tang.
\newblock Convolutional neural network models of v1 responses to complex
  patterns.
\newblock \emph{Journal of computational neuroscience}, 46\penalty0
  (1):\penalty0 33--54, 2019.

\bibitem[Benda et~al.(2007)Benda, Gollisch, Machens, and Herz]{benda2007}
Jan Benda, Tim Gollisch, Christian~K Machens, and Andreas~VM Herz.
\newblock From response to stimulus: adaptive sampling in sensory physiology.
\newblock \emph{Current opinion in neurobiology}, 17\penalty0 (4):\penalty0
  430--436, 2007.

\bibitem[DiMattina and Zhang(2011)]{dimattina2011}
Christopher DiMattina and Kechen Zhang.
\newblock Active data collection for efficient estimation and comparison of
  nonlinear neural models.
\newblock \emph{Neural computation}, 23\penalty0 (9):\penalty0 2242--2288,
  2011.

\bibitem[Cowley et~al.(2017)Cowley, Williamson, Acar, Smith, and
  Yu]{cowley2017}
Benjamin~R Cowley, Ryan~C Williamson, Katerina Acar, Matthew~A Smith, and
  Byron~M Yu.
\newblock Adaptive stimulus selection for optimizing neural population
  responses.
\newblock In \emph{Advances in neural information processing systems}, pages
  1396--1406, 2017.

\bibitem[Ponce et~al.(2019)Ponce, Xiao, Schade, Hartmann, Kreiman, and
  Livingstone]{ponce2019}
Carlos~R Ponce, Will Xiao, Peter~F Schade, Till~S Hartmann, Gabriel Kreiman,
  and Margaret~S Livingstone.
\newblock Evolving images for visual neurons using a deep generative network
  reveals coding principles and neuronal preferences.
\newblock \emph{Cell}, 177\penalty0 (4):\penalty0 999--1009, 2019.

\bibitem[Walker et~al.(2019)Walker, Sinz, Cobos, Muhammad, Froudarakis, Fahey,
  Ecker, Reimer, Pitkow, and Tolias]{walker2019}
Edgar~Y Walker, Fabian~H Sinz, Erick Cobos, Taliah Muhammad, Emmanouil
  Froudarakis, Paul~G Fahey, Alexander~S Ecker, Jacob Reimer, Xaq Pitkow, and
  Andreas~S Tolias.
\newblock Inception loops discover what excites neurons most using deep
  predictive models.
\newblock \emph{Nature neuroscience}, 22\penalty0 (12):\penalty0 2060--2065,
  2019.

\bibitem[Bashivan et~al.(2019)Bashivan, Kar, and DiCarlo]{bashivan2019}
Pouya Bashivan, Kohitij Kar, and James~J DiCarlo.
\newblock Neural population control via deep image synthesis.
\newblock \emph{Science}, 364\penalty0 (6439):\penalty0 eaav9436, 2019.

\bibitem[Morcos et~al.(2018)Morcos, Raghu, and Bengio]{morcos2018}
Ari Morcos, Maithra Raghu, and Samy Bengio.
\newblock Insights on representational similarity in neural networks with
  canonical correlation.
\newblock In \emph{Advances in Neural Information Processing Systems}, pages
  5727--5736, 2018.

\bibitem[Kornblith et~al.(2019)Kornblith, Norouzi, Lee, and
  Hinton]{kornblith2019}
Simon Kornblith, Mohammad Norouzi, Honglak Lee, and Geoffrey Hinton.
\newblock Similarity of neural network representations revisited.
\newblock \emph{arXiv preprint arXiv:1905.00414}, 2019.

\bibitem[Lu et~al.(2018)Lu, Chen, Pillow, Ramadge, Norman, and Hasson]{lu2018}
Qihong Lu, Po-Hsuan Chen, Jonathan~W Pillow, Peter~J Ramadge, Kenneth~A Norman,
  and Uri Hasson.
\newblock Shared representational geometry across neural networks.
\newblock \emph{arXiv preprint arXiv:1811.11684}, 2018.

\bibitem[Pukelsheim(2006)]{pukelsheim2006}
Friedrich Pukelsheim.
\newblock \emph{Optimal design of experiments}.
\newblock SIAM, 2006.

\bibitem[Jaakkola(2006)]{jaakkola2006}
Tommi Jaakkola.
\newblock Course materials for 6.867 machine learning, fall 2006, 2006.

\bibitem[He et~al.(2016)He, Zhang, Ren, and Sun]{he2016}
Kaiming He, Xiangyu Zhang, Shaoqing Ren, and Jian Sun.
\newblock Deep residual learning for image recognition.
\newblock In \emph{Proceedings of the IEEE conference on computer vision and
  pattern recognition}, pages 770--778, 2016.

\bibitem[Yamins and DiCarlo(2016)]{yamins2016}
Daniel~LK Yamins and James~J DiCarlo.
\newblock Using goal-driven deep learning models to understand sensory cortex.
\newblock \emph{Nature neuroscience}, 19\penalty0 (3):\penalty0 356, 2016.

\bibitem[Ji et~al.(2016)Ji, Freeman, and Smith]{ji2016}
Na~Ji, Jeremy Freeman, and Spencer~L Smith.
\newblock Technologies for imaging neural activity in large volumes.
\newblock \emph{Nature neuroscience}, 19\penalty0 (9):\penalty0 1154, 2016.

\bibitem[Sheintuch et~al.(2017)Sheintuch, Rubin, Brande-Eilat, Geva, Sadeh,
  Pinchasof, and Ziv]{sheintuch2017}
Liron Sheintuch, Alon Rubin, Noa Brande-Eilat, Nitzan Geva, Noa Sadeh,
  Or~Pinchasof, and Yaniv Ziv.
\newblock Tracking the same neurons across multiple days in ca2+ imaging data.
\newblock \emph{Cell reports}, 21\penalty0 (4):\penalty0 1102--1115, 2017.

\bibitem[Nonnenmacher et~al.(2017)Nonnenmacher, Turaga, and
  Macke]{nonnenmacher2017}
Marcel Nonnenmacher, Srinivas~C Turaga, and Jakob~H Macke.
\newblock Extracting low-dimensional dynamics from multiple large-scale neural
  population recordings by learning to predict correlations.
\newblock In \emph{Advances in Neural Information Processing Systems}, pages
  5702--5712, 2017.

\bibitem[Degenhart et~al.(2020)Degenhart, Bishop, Oby, Tyler-Kabara, Chase,
  Batista, and Byron]{degenhart2020}
Alan~D Degenhart, William~E Bishop, Emily~R Oby, Elizabeth~C Tyler-Kabara,
  Steven~M Chase, Aaron~P Batista, and M~Yu Byron.
\newblock Stabilization of a brain--computer interface via the alignment of
  low-dimensional spaces of neural activity.
\newblock \emph{Nature Biomedical Engineering}, pages 1--14, 2020.

\bibitem[Simonyan and Zisserman(2014)]{simonyan2014}
Karen Simonyan and Andrew Zisserman.
\newblock Very deep convolutional networks for large-scale image recognition.
\newblock \emph{arXiv preprint arXiv:1409.1556}, 2014.

\bibitem[Szegedy et~al.(2016)Szegedy, Vanhoucke, Ioffe, Shlens, and
  Wojna]{szegedy2016}
Christian Szegedy, Vincent Vanhoucke, Sergey Ioffe, Jon Shlens, and Zbigniew
  Wojna.
\newblock Rethinking the inception architecture for computer vision.
\newblock In \emph{Proceedings of the IEEE conference on computer vision and
  pattern recognition}, pages 2818--2826, 2016.

\bibitem[Huang et~al.(2017{\natexlab{a}})Huang, Liu, Van Der~Maaten, and
  Weinberger]{huang2017}
Gao Huang, Zhuang Liu, Laurens Van Der~Maaten, and Kilian~Q Weinberger.
\newblock Densely connected convolutional networks.
\newblock In \emph{Proceedings of the IEEE conference on computer vision and
  pattern recognition}, pages 4700--4708, 2017{\natexlab{a}}.

\bibitem[Settles(2012)]{settles2012}
Burr Settles.
\newblock Active learning. morgan claypool.
\newblock \emph{Synthesis Lectures on AI and ML}, 2012.

\bibitem[Huang et~al.(2017{\natexlab{b}})Huang, Li, Pleiss, Liu, Hopcroft, and
  Weinberger]{huang2017snapshot}
Gao Huang, Yixuan Li, Geoff Pleiss, Zhuang Liu, John~E Hopcroft, and Kilian~Q
  Weinberger.
\newblock Snapshot ensembles: Train 1, get m for free.
\newblock \emph{arXiv preprint arXiv:1704.00109}, 2017{\natexlab{b}}.

\bibitem[Tran et~al.(2019)Tran, Do, Reid, and Carneiro]{tran2019}
Toan Tran, Thanh-Toan Do, Ian Reid, and Gustavo Carneiro.
\newblock Bayesian generative active deep learning.
\newblock \emph{arXiv preprint arXiv:1904.11643}, 2019.

\bibitem[Sinha et~al.(2019)Sinha, Ebrahimi, and Darrell]{sinha2019}
Samarth Sinha, Sayna Ebrahimi, and Trevor Darrell.
\newblock Variational adversarial active learning.
\newblock In \emph{Proceedings of the IEEE International Conference on Computer
  Vision}, pages 5972--5981, 2019.

\bibitem[Sener and Savarese(2017)]{sener2017}
Ozan Sener and Silvio Savarese.
\newblock Active learning for convolutional neural networks: A core-set
  approach.
\newblock \emph{arXiv preprint arXiv:1708.00489}, 2017.

\bibitem[Beluch et~al.(2018)Beluch, Genewein, N{\"u}rnberger, and
  K{\"o}hler]{beluch2018}
William~H Beluch, Tim Genewein, Andreas N{\"u}rnberger, and Jan~M K{\"o}hler.
\newblock The power of ensembles for active learning in image classification.
\newblock In \emph{Proceedings of the IEEE Conference on Computer Vision and
  Pattern Recognition}, pages 9368--9377, 2018.

\bibitem[Gal et~al.(2017)Gal, Islam, and Ghahramani]{gal2017}
Yarin Gal, Riashat Islam, and Zoubin Ghahramani.
\newblock Deep bayesian active learning with image data.
\newblock In \emph{Proceedings of the 34th International Conference on Machine
  Learning-Volume 70}, pages 1183--1192. JMLR. org, 2017.

\bibitem[Kirsch et~al.(2019)Kirsch, van Amersfoort, and Gal]{kirsch2019}
Andreas Kirsch, Joost van Amersfoort, and Yarin Gal.
\newblock Batchbald: Efficient and diverse batch acquisition for deep bayesian
  active learning.
\newblock In \emph{Advances in Neural Information Processing Systems}, pages
  7024--7035, 2019.

\bibitem[Yoo and Kweon(2019)]{yoo2019}
Donggeun Yoo and In~So Kweon.
\newblock Learning loss for active learning.
\newblock In \emph{Proceedings of the IEEE Conference on Computer Vision and
  Pattern Recognition}, pages 93--102, 2019.

\bibitem[Ducoffe and Precioso(2018)]{ducoffe2018}
Melanie Ducoffe and Frederic Precioso.
\newblock Adversarial active learning for deep networks: a margin based
  approach.
\newblock \emph{arXiv preprint arXiv:1802.09841}, 2018.

\bibitem[Munjal et~al.(2020)Munjal, Hayat, Hayat, Sourati, and
  Khan]{munjal2020}
Prateek Munjal, Nasir Hayat, Munawar Hayat, Jamshid Sourati, and Shadab Khan.
\newblock Towards robust and reproducible active learning using neural
  networks.
\newblock \emph{arXiv}, pages arXiv--2002, 2020.

\bibitem[Ash et~al.(2019)Ash, Zhang, Krishnamurthy, Langford, and
  Agarwal]{ash2019}
Jordan~T Ash, Chicheng Zhang, Akshay Krishnamurthy, John Langford, and Alekh
  Agarwal.
\newblock Deep batch active learning by diverse, uncertain gradient lower
  bounds.
\newblock \emph{arXiv preprint arXiv:1906.03671}, 2019.

\bibitem[Tsymbalov et~al.(2018)Tsymbalov, Panov, and Shapeev]{tsymbalov2018}
Evgenii Tsymbalov, Maxim Panov, and Alexander Shapeev.
\newblock Dropout-based active learning for regression.
\newblock In \emph{International Conference on Analysis of Images, Social
  Networks and Texts}, pages 247--258. Springer, 2018.

\bibitem[Dosovitskiy and Brox(2016)]{dosovitskiy2016}
Alexey Dosovitskiy and Thomas Brox.
\newblock Generating images with perceptual similarity metrics based on deep
  networks.
\newblock In \emph{Advances in neural information processing systems}, pages
  658--666, 2016.

\bibitem[Sohn et~al.(2020)Sohn, Berthelot, Li, Zhang, Carlini, Cubuk, Kurakin,
  Zhang, and Raffel]{sohn2020}
Kihyuk Sohn, David Berthelot, Chun-Liang Li, Zizhao Zhang, Nicholas Carlini,
  Ekin~D Cubuk, Alex Kurakin, Han Zhang, and Colin Raffel.
\newblock Fixmatch: Simplifying semi-supervised learning with consistency and
  confidence.
\newblock \emph{arXiv preprint arXiv:2001.07685}, 2020.

\bibitem[Perez and Wang(2017)]{perez2017}
Luis Perez and Jason Wang.
\newblock The effectiveness of data augmentation in image classification using
  deep learning.
\newblock \emph{arXiv preprint arXiv:1712.04621}, 2017.

\bibitem[Shorten and Khoshgoftaar(2019)]{shorten2019}
Connor Shorten and Taghi~M Khoshgoftaar.
\newblock A survey on image data augmentation for deep learning.
\newblock \emph{Journal of Big Data}, 6\penalty0 (1):\penalty0 60, 2019.

\bibitem[Moreno-Bote et~al.(2014)Moreno-Bote, Beck, Kanitscheider, Pitkow,
  Latham, and Pouget]{moreno2014}
Rub{\'e}n Moreno-Bote, Jeffrey Beck, Ingmar Kanitscheider, Xaq Pitkow, Peter
  Latham, and Alexandre Pouget.
\newblock Information-limiting correlations.
\newblock \emph{Nature neuroscience}, 17\penalty0 (10):\penalty0 1410, 2014.

\bibitem[Parthasarathy et~al.(2017)Parthasarathy, Batty, Falcon, Rutten,
  Rajpal, Chichilnisky, and Paninski]{parthasarathy2017}
Nikhil Parthasarathy, Eleanor Batty, William Falcon, Thomas Rutten, Mohit
  Rajpal, EJ~Chichilnisky, and Liam Paninski.
\newblock Neural networks for efficient bayesian decoding of natural images
  from retinal neurons.
\newblock In \emph{Advances in Neural Information Processing Systems}, pages
  6434--6445, 2017.

\bibitem[Xiao and Kreiman(2019)]{xiao2019}
Will Xiao and Gabriel Kreiman.
\newblock Gradient-free activation maximization for identifying effective
  stimuli.
\newblock \emph{arXiv preprint arXiv:1905.00378}, 2019.

\bibitem[Lewi et~al.(2009)Lewi, Butera, and Paninski]{lewi2009}
Jeremy Lewi, Robert Butera, and Liam Paninski.
\newblock Sequential optimal design of neurophysiology experiments.
\newblock \emph{Neural computation}, 21\penalty0 (3):\penalty0 619--687, 2009.

\bibitem[Park et~al.(2011)Park, Horwitz, and Pillow]{park2011}
Mijung Park, Greg Horwitz, and Jonathan~W Pillow.
\newblock Active learning of neural response functions with gaussian processes.
\newblock In \emph{Advances in neural information processing systems}, pages
  2043--2051, 2011.

\bibitem[Sharpee(2013)]{sharpee2013}
Tatyana~O Sharpee.
\newblock Computational identification of receptive fields.
\newblock \emph{Annual review of neuroscience}, 36:\penalty0 103--120, 2013.

\bibitem[Montijn et~al.(2019)Montijn, Liu, Aschner, Kohn, Latham, and
  Pouget]{montijn2019}
Jorrit~Steven Montijn, Rex~G Liu, Amir Aschner, Adam Kohn, Peter~E Latham, and
  Alexandre Pouget.
\newblock Strong information-limiting correlations in early visual areas.
\newblock \emph{bioRxiv}, page 842724, 2019.

\bibitem[Brackbill et~al.(2020)Brackbill, Rhoades, Kling, Shah, Sher, Litke,
  and Chichilnisky]{brackbill2020}
Nora Brackbill, Colleen Rhoades, Alexandra Kling, Nishal~P Shah, Alexander
  Sher, Alan~M Litke, and EJ~Chichilnisky.
\newblock Reconstruction of natural images from responses of primate retinal
  ganglion cells.
\newblock \emph{bioRxiv}, 2020.

\bibitem[Chollet et~al.(2015)]{chollet2015}
Fran\c{c}ois Chollet et~al.
\newblock Keras.
\newblock \url{https://keras.io}, 2015.

\bibitem[Abadi et~al.(2016)Abadi, Barham, Chen, Chen, Davis, Dean, Devin,
  Ghemawat, Irving, Isard, et~al.]{abadi2016}
Mart{\'\i}n Abadi, Paul Barham, Jianmin Chen, Zhifeng Chen, Andy Davis, Jeffrey
  Dean, Matthieu Devin, Sanjay Ghemawat, Geoffrey Irving, Michael Isard, et~al.
\newblock Tensorflow: A system for large-scale machine learning.
\newblock In \emph{12th $\{$USENIX$\}$ Symposium on Operating Systems Design
  and Implementation ($\{$OSDI$\}$ 16)}, pages 265--283, 2016.

\bibitem[Thomee et~al.(2016)Thomee, Shamma, Friedland, Elizalde, Ni, Poland,
  Borth, and Li]{thomee2016}
Bart Thomee, David~A Shamma, Gerald Friedland, Benjamin Elizalde, Karl Ni,
  Douglas Poland, Damian Borth, and Li-Jia Li.
\newblock Yfcc100m: The new data in multimedia research.
\newblock \emph{Communications of the ACM}, 59\penalty0 (2):\penalty0 64--73,
  2016.

\bibitem[Deng et~al.(2009)Deng, Dong, Socher, Li, Li, and Fei-Fei]{deng2009}
Jia Deng, Wei Dong, Richard Socher, Li-Jia Li, Kai Li, and Li~Fei-Fei.
\newblock Imagenet: A large-scale hierarchical image database.
\newblock In \emph{2009 IEEE conference on computer vision and pattern
  recognition}, pages 248--255. Ieee, 2009.

\bibitem[Han et~al.(2015)Han, Pool, Tran, and Dally]{han2015}
Song Han, Jeff Pool, John Tran, and William Dally.
\newblock Learning both weights and connections for efficient neural network.
\newblock In \emph{Advances in neural information processing systems}, pages
  1135--1143, 2015.

\bibitem[Frankle and Carbin(2018)]{frankle2018}
Jonathan Frankle and Michael Carbin.
\newblock The lottery ticket hypothesis: Finding sparse, trainable neural
  networks.
\newblock \emph{arXiv preprint arXiv:1803.03635}, 2018.

\bibitem[Ioffe and Szegedy(2015)]{ioffe2015}
Sergey Ioffe and Christian Szegedy.
\newblock Batch normalization: Accelerating deep network training by reducing
  internal covariate shift.
\newblock \emph{arXiv preprint arXiv:1502.03167}, 2015.

\bibitem[Goodfellow et~al.(2014)Goodfellow, Shlens, and
  Szegedy]{goodfellow2014}
Ian~J Goodfellow, Jonathon Shlens, and Christian Szegedy.
\newblock Explaining and harnessing adversarial examples.
\newblock \emph{arXiv preprint arXiv:1412.6572}, 2014.

\bibitem[Davies(2004)]{davies2004}
E~Roy Davies.
\newblock \emph{Machine vision: theory, algorithms, practicalities}.
\newblock Elsevier, 2004.

\bibitem[Cubuk et~al.(2019)Cubuk, Zoph, Shlens, and Le]{cubuk2019}
Ekin~D Cubuk, Barret Zoph, Jonathon Shlens, and Quoc~V Le.
\newblock Randaugment: Practical automated data augmentation with a reduced
  search space.
\newblock \emph{arXiv preprint arXiv:1909.13719}, 2019.

\end{thebibliography}
\end{document}